\pdfoutput=1
\documentclass[11pt]{article}
\usepackage[inkscapelatex=false]{svg}
\usepackage{enumitem}
\usepackage{ACL2023}
\usepackage{tabularray}
\usepackage{times}
\usepackage{latexsym}
\usepackage{booktabs,tabularx}
\newcolumntype{Y}{>{\raggedright\arraybackslash}X} 
\usepackage[T1]{fontenc}
\usepackage[utf8]{inputenc}
\usepackage{color,soul}
\usepackage{microtype}
\usepackage[toc,page]{appendix}
\usepackage{inconsolata}
\usepackage{multirow}
\usepackage{graphicx}
\usepackage{tcolorbox} 
\usepackage{inconsolata}
\usepackage{booktabs}
\usepackage{dingbat}
\usepackage{xcolor,colortbl}
\definecolor{lightgray}{gray}{0.9}
\setlist[itemize,1]{leftmargin=\dimexpr 15pt}

\newtcolorbox{mybox}[3][]
{
  colframe = #2!25,
  colback  = #2!10,
  coltitle = #2!20!black,  
  title    = {#3},
  #1,
}

\title{Decoding the Narratives: Analyzing Personal Drug Experiences Shared on Reddit}
\author{Layla Bouzoubaa, Elham Aghakhani, Max Song, Minh Trinh, Rezvaneh Rezapour \\
Drexel University\\
  \texttt{\{lb3338, ea664, ms5526, mqt32, shadi.rezapour\}@drexel.edu} \\}

\begin{document}
\maketitle
\begin{abstract}

Online communities such as drug-related subreddits serve as safe spaces for people who use drugs (PWUD), fostering discussions on substance use experiences, harm reduction, and addiction recovery. Users' shared narratives on these forums provide insights into the likelihood of developing a substance use disorder (SUD) and recovery potential. Our study aims to develop a multi-level, multi-label classification model to analyze online user-generated texts about substance use experiences. For this purpose, we first introduce a novel taxonomy to assess the nature of posts, including their intended connections (Inquisition or Disclosure), subjects (e.g., Recovery, Dependency), and specific objectives (e.g., Relapse, Quality, Safety). Using various multi-label classification algorithms on a set of annotated data, we show that GPT-4, when prompted with instructions, definitions, and examples, outperformed all other models. We apply this model to label an additional 1,000 posts and analyze the categories of linguistic expression used within posts in each class. Our analysis shows that topics such as Safety, Combination of Substances, and Mental Health see more disclosure, while discussions about physiological Effects focus on harm reduction. Our work enriches the understanding of PWUD's experiences and informs the broader knowledge base on SUD and drug use.

\end{abstract}
    
\section{Introduction}
\textbf{\textcolor{blue}{Warning:}} \textcolor{blue} {This paper includes language and content that may be offensive or triggering.}

For people who use drugs (PWUD), social platforms like Reddit serve as invaluable spaces for open discussion and community support. Such platforms enable PWUD to engage in conversations and share experiences, facilitated by the anonymity and community solidarity that Reddit provides (Figure \ref{fig:e1}). Despite Reddit's ability to foster connection through shared experiences \cite{bouzoubaa2023exploring,choudhury_mental_2014}, the perspectives of PWUD are frequently marginalized in important decision-making processes, particularly for those who depend exclusively on online communities due to the significant stigma associated with seeking help from traditional services \cite{volkow_choosing_2021}.

\begin{figure}
\centering
\includegraphics[width=\linewidth]{chatboxc.png}
% \includesvg[width=0.95\columnwidth]{chat}
\caption{Examples of user-generated posts on drug-related subreddits}
\label{fig:e1}
\vspace{-0.5cm}
\end{figure}

Analyzing how PWUD communicate about their experiences on online platforms offers important insights into their narratives, highlighting their information-seeking behaviors and the diverse needs specific to this population \cite{valdez_computational_2022}, revealing the real-life challenges and perspectives of individuals dealing with Substance Use Disorders (SUD) \cite{brown2019achieving, Bouzoubaa_Rezapour_2024}. 
% This nuanced understanding is crucial for devising health interventions that are both supportive and capable of navigating the intricacies associated with drug use within both digital and physical community settings.
Existing literature in this domain primarily focuses on classifying mental health experiences like depression \cite{rijen_data-driven_2019} or drug-related events like abuse \cite{al-garadi_text_2021} or overdose risk \cite{garg_detecting_2021}. Personal lived experiences of PWUD remain largely unexplored, highlighting the need for effective methodologies to understand and analyze diverse drug experiences shared online. 

Our study aims to bridge this gap by exploring a range of substance use experiences shared online, from recovery to general discussions, through the development of a taxonomy and model for classifying these narratives on Reddit. 
Insights from this study will not only expand our understanding of how PWUD navigate information seeking and support mechanisms, thereby laying the groundwork for harm reduction strategies and more nuanced, effective interventions, but also help in identifying and understanding the diverse experiences and characteristics of individuals most impacted by drug misuse \cite{strapparava_computational_2017, yang_can_2023}.

To understand the experiences of PWUD, we first developed a fine-grained taxonomy of online drug experiences comprising three levels: Connection (which highlights the information-seeking and sharing behaviors of users), Subject (lived SUD-related experiences such as Dependency and Recovery), and Objectives (detailing the aspects shared within posts related to SUD experiences). After testing and evaluating our new taxonomy, we annotated posts and developed multi-level and multi-label classification models using different baseline and state-of-the-art approaches. These models were used to categorize different types of personal drug experiences shared in posts from four prominent drug-related subreddits \textit{r/cocaine}, \textit{r/opiates}, \textit{r/stims}, \textit{r/benzodiazepines}. These subreddits were selected as they are the largest subreddits that represent substances identified as most commonly abused \cite{nida_commonly_used_drugs_charts}. 
The results of our analysis show that GPT-4 outperformed other models and more accurately labeled classes across each of the three levels. Further exploration of the labeled posts showed that online posts are more inquisitive in nature and discuss themes relating to desired or undesired effects of the substance and/or how to consume them. Especially in the context of Recovery-related posts, users emphasize themes of Nurturant support, Relapse, and Safety. Moreover, our results show that discussions around Dependency frequently cover the Effects and Methods of Ingestion. Psycholinguistic analysis revealed that posts that contain topics such as Safety, Combination of Substances, and Effects tend to share personal experiences and exhibit a higher prevalence of language indicative of harm reduction efforts, as well as personal disclosures on family and social support systems. 

Our study makes several contributions. Firstly, it introduces a new taxonomy for the classification of personal drug experiences. Additionally, we have developed a human-annotated dataset comprising 500 Reddit posts related to drug use, which showcases the wide range of personal drug experiences discussed in user-generated content. We also demonstrate the capability for automatic classification of personal drug experiences across three levels and multiple classes. Lastly, our work includes an analysis of the narratives surrounding substance use, SUD, and recovery based on self-disclosed user experiences. This work enriches the understanding of PWUD's experiences and informs the broader knowledge base on SUD and drug use.

\section{Related Work}
\subsection{Exploring Personal SUD Narratives}

Within the two drug-related subreddits, \textit{r/trees} and \textit{r/opiates}, \newcite{Costello2017} utilized hermeneutic content analysis to categorize forum posts into eight distinct groups: disclosure, instruction, drug culture, community norms, moralizing, legality, and banter and identified three primary motives for PWUD to share information: to offer advice, seek information from others, and provide context for illicit disclosures. \newcite{wombacher_social_2020} applied content analysis to the well-known drug subreddit, \textit{r/Drugs}, to identify the types of social support exchanged among active substance users and found that the majority of the support was action-facilitating, aimed at safer drug use, with emotional support also being significant. 

Recent studies used NLP and machine learning to analyze drug-related discussions on social media, offering insights into user behavior and content shared. \newcite{balani_detecting_2015} analyzed over 30,000 posts from mental health-related subreddits, and found a significant amount of self-disclosure among users, with those engaging more intensely showing longer activity on the platform. 
\newcite{strapparava_computational_2017} used mental health forum posts to classify DSM-5 categories via zero-shot learning, employing n-grams and LDA topics. Incorporating slang improved accuracy by reducing false alarms by 17\%, and domain knowledge enhanced recall.
\newcite{varma_few_2022} applied few-shot learning models to identify suicide risk on Reddit and found that few-shot learning with outlier removal and Support Vector Machine classifier yields better accuracy in detecting suicide risk.

Our study extends prior work by incorporating insights, taxonomies, and methods from existing research on online communications. We developed a refined and extensive taxonomy for analyzing user-generated texts, alongside a codebook for human annotation. This new taxonomy aims to provide a theoretical foundation, ensuring accuracy and consistency in annotating user interactions.

\subsection{LLM-based Information Extraction and Annotation}
In recent years, the rapid advancements of LLMs such as GPT models, LLaMA, OPT, and BLOOM, have facilitated various tasks in NLP \cite{wei2022emergent}, demonstrating great performance across a wide range of NLP tasks such as question answering \cite{trivedi2022interleaving}, named-entity recognition \cite{wang2023gpt}, dialogue \cite{thoppilan2022lamda}, translation \cite{peng2023towards}, and emotion analysis \cite{lei2023instructerc}, often outperforming other models in zero-shot and few-shot contexts. LLMs' in-context learning (ICL) and text classification capabilities enable generating prompt-based textual responses, often with minimal examples \cite{wei2022chain,sun2023text}. Their proficiency in mimicking human text and labeling facilitates scalable NLP tasks in information retrieval with context sensitivity \cite{li_evaluating_2023}. 
\newcite{brown2020language} demonstrated a few-shot classification method with GPT-3 for various NLP tasks, using text-based task definitions and demonstrations. The model showcased proficiency in on-the-fly reasoning and domain adaptation tasks like word unscrambling, novel word usage, and 3-digit arithmetic.

LLMs have been applied in human-in-the-loop and co-annotation methods \cite{li_coannotating_2023}. Cost efficiency is improved by using consensus methods between human and AI outputs \cite{chaganty2018price, zhang2021creating}. \newcite{kang2023distill} explored combining LLM distillation with manual annotation, while \newcite{wang2021want} looked into active labeling via logit outputs. While co-annotation balances quality and cost by leveraging uncertainty measures and evaluation thresholds \cite{li_coannotating_2023}, it faces challenges, as LLMs inherently struggle with generating structured abstract meaning, necessitating additional adaptation techniques \cite{ettinger_you_2023}.

The use of LLMs in the healthcare domain has shown promise in expanding the capacity of tasks such as summarization of patients' health records or question-answering (chatbot) \cite{liu2023deid,nov2023putting}. For instance, \newcite{garg_detecting_2021} used LLMs to detect patient deterioration from electronic health records, demonstrating the potential of LLMs to accurately identify critical events. Similarly, \newcite{rijen_data-driven_2019} utilized LLMs to classify online health forum posts, demonstrating their ability to handle free-text data. Drawing on insights from previous studies, we employed a range of models, including LLM, to extract domain-specific information and classify personal drug experiences in user-generated content. 
% data collection process should go before taxonomy i think
\section{Data \& Taxonomy Development} \label{sec:data}

\subsection{Data Collection}
Data for this study was obtained using the Reddit API and the Python for Reddit API Wrapper (PRAW).\footnote{\url{https://praw.readthedocs.io/en/stable}} We collected posts from four popular drug subreddits (\textit{r/opiate}s, \textit{r/benzodiazepines}, \textit{r/stims}, and \textit{r/cocaine}) posted between 2017 and 2022, resulting in collecting 267,748 posts. These subreddits were selected because they are the largest within their respective classes of commonly abused substances \cite{nida_commonly_used_drugs_charts}. 
We developed the taxonomy and training/test sets by randomly sampling around 1,600 posts. Initially, 100 posts were divided into four sets of 25 for the preliminary labeling and taxonomy development. We then annotated 500 posts for model training and testing. An additional 1,000 posts were then selected for labeling using the best-performing model.

\begin{figure*}[ht]
    \centering
    \includegraphics[width = 0.85\textwidth]{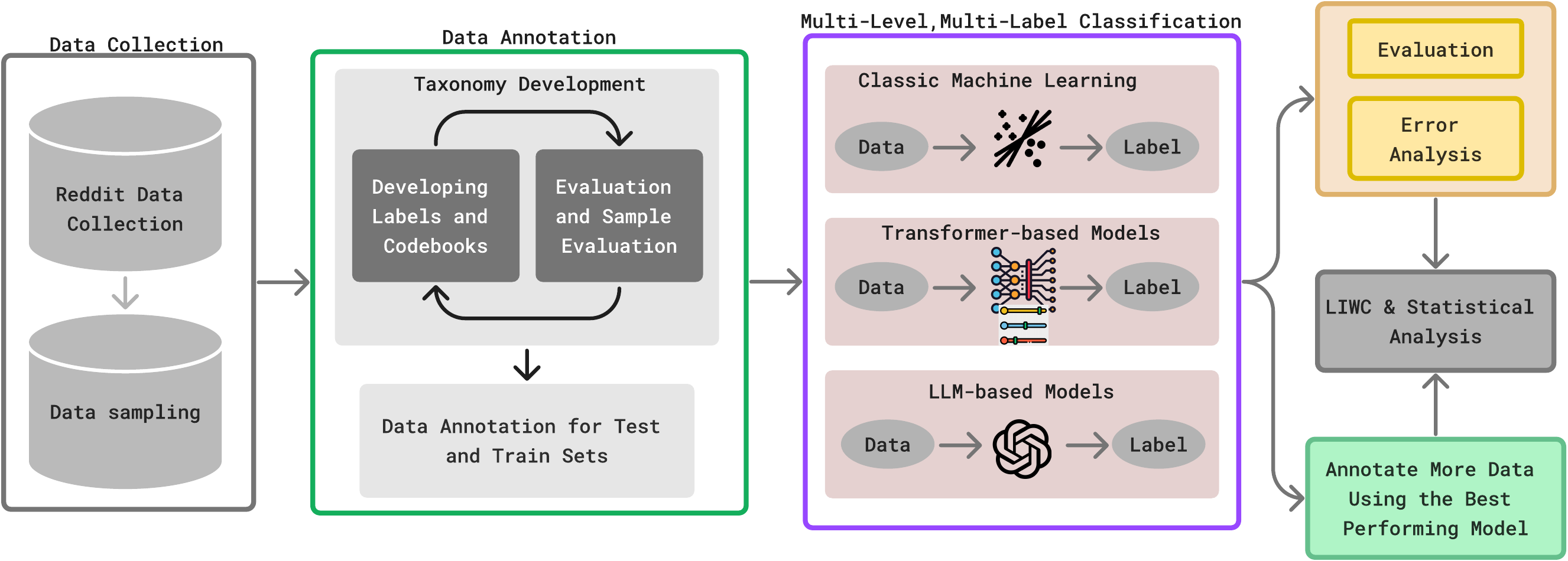}
    \caption{Classification and data analysis pipeline.}
    \label{fig:pipeline}
    
\end{figure*}

\subsection{Taxonomy of Lived Experiences Online}
To understand the nuanced experiences of PWUD, we develop a domain-specific taxonomy derived from analyzing user-generated texts to distill narratives and offer insights into the personal and social aspects of drug use on social media. To develop this taxonomy and a codebook for annotation, we employed a hybrid deductive-inductive approach. First, using the existing literature in the field of drug use and social media, we established a framework for understanding key themes and dimensions related to personal drug experiences. This deductive phase involved developing a set of codes based on concepts from the literature, such as the type or objectives of posts (giving or receiving advice) \cite{balani_detecting_2015, valdez_computational_2022}, indication of social support \cite{gauthier_i_2022}, and recovery- or withdrawal-related discourse \cite{dagostino_social_2017}. This approach allowed us to create a comprehensive codebook tailored to our data and grounded in existing theories and research. 
Next, we applied these deductive codes to a sample of 25 Reddit posts. During coding, we asked our annotators (graduate and undergraduate students with a variety of experiences in health and computing) to identify new themes and concepts that were not covered by our initial codes. This resulted in the extraction of new concepts that were later added to the codebook as inductive codes. After coding a new sample of 25 posts, we reviewed and revised the codebook as necessary. This iterative process was repeated three times before we finalized the taxonomy and codebook, ensuring that it was well-suited to our data. In each iteration, annotators labeled a new set of posts in addition to those we labeled before to assess confidence in theme creation. 

The final codebook consists of three levels: (1) Connection, (2) Subject, and (3) Objective. The type of Connection determines the overarching intent of the post, whether the user seeks to gain (Inquisition) or give information (Disclosure). Subject refers to the essence of the lived experience mentioned, particularly around Dependency, Recovery, and Other (typically recreational) experiences. 
While we labeled recovery and dependency-based discourse separately, we ensured that other circumstances were captured in our taxonomy. For example, posts referring to medicinal experiences (e.g., pain or anxiety) or an experience non-indicative of abuse were categorized as Other.
Finally, the Objective captures the fine-grained topics discussed in the posts. For example, a post with a Recovery subject could include information on Safety, Quality, and Overdose. We consulted with domain experts in health to validate our new taxonomy, ensuring its efficiency and usefulness for professionals in the field.
Table \ref{tab:definition} in Appendix \ref{appendix:definitions} provides an overview of the taxonomy for lived drug experiences, featuring definitions and examples for each code.

\vspace{-0.35cm}
\subsection{Annotation Process}
After finalizing the codebook, three of the authors annotated two sets of 50 randomly selected posts. The agreement between each set of annotations was calculated ($k$ = .78 for Connection, and $k$ = .51 for Subject)\footnote{Due to the complexity of the Objective level, despite multiple training sessions, there was still no substantial agreement between annotators across all 13 classes. 
With additional training, it may be possible to achieve higher inter-coder reliability.}, and any posts with disagreements were thoroughly discussed to establish a set of 100 mutually agreed-upon annotated posts. This dataset was used to test and evaluate the classification models.
Once the annotators demonstrated a good understanding of the codebook, they annotated the 400 remaining posts. Regular check-ins were conducted between the annotators to ensure quality and consistency in using the codebook. %The 400 annotated posts were then used to train a set of classification models, as discussed in the following sections.

\section{Multi-Level Lived Experience Detection}
%\subsection{Multi-Label Classification}
We used 400 annotated posts to train a series of multi-label models, each customized for different levels of classification granularity: Connection, Subject, and Objective. Minimal pre-processing was applied to maintain data integrity, which included converting text to lowercase, removing URLs, expanding contractions, and eliminating stopwords. We evaluate the models' performance on the test set consisting of 100 posts, using precision, recall, and F1 score metrics. Figure \ref{fig:pipeline} shows the overall pipeline of this study. 
\vspace{-0.25cm}
\paragraph{Baseline Models.} We implemented four baseline machine learning models to benchmark our drug experience classification system. We chose classic, feature-based algorithms for their interpretability and widespread use: Logistic Regression (LogR), Random Forest (RF), Support Vector Machines (SVM), and K-Nearest Neighbors (KNN). To provide a comprehensive evaluation, we trained and tested each model with two distinct feature sets: TF-IDF vectors and BERT \cite{Devlin_Chang_Lee_Toutanova_2019} pre-trained [CLS] token embedding. This approach allowed us to understand the impact of feature representation on model performance.
In addition, we integrated OpenAI's \texttt{ada-002} architecture into our approach. Leveraging the capabilities of these LLM, we generated high-level features and trained our selected classic models.%, namely RF, SVM, LogR, and KNN. 

\vspace{-0.25cm}
\paragraph{Transformer-based Models. } We used pre-trained language models from Hugging Face \cite{wolf2020transformers} to extract rich contextual features. Specifically, BERT-base-uncased \cite{Devlin_Chang_Lee_Toutanova_2019}, RoBERTa-base \cite{Liu_2019}, DeBERTa-base \cite{He_Liu_Gao_Chen_2021}, and BioBERT-v1.1 \cite{Lee_Yoon_Kim_Kim_Kim_So_Kang_2020} were used. All models were fine-tuned and trained using a cross entropy loss function for 5 epochs and a learning rate of $2e-5$. During training, we optimized the model with AdamW optimizer\cite{loshchilov2018decoupled}.

Recognizing the limitations of large-scale annotation, we explored few-shot learning techniques. We employed SetFit \cite{Tunstall_2022}, one of the most efficient few-shot architectures, leveraging SentBERT \cite{Reimers_Gurevych_2019} features. This choice aligns with current research advocating for few-shot models in situations where extensive annotation is impractical or costly \cite{schick_exploiting_2021}. To optimize performance, we fine-tuned the model using the cosine similarity loss function, for 2 epochs with a learning rate of $2e-5$ and a batch size of 8.

\begin{figure}[t]
    \centering
    \includegraphics[width=\columnwidth]{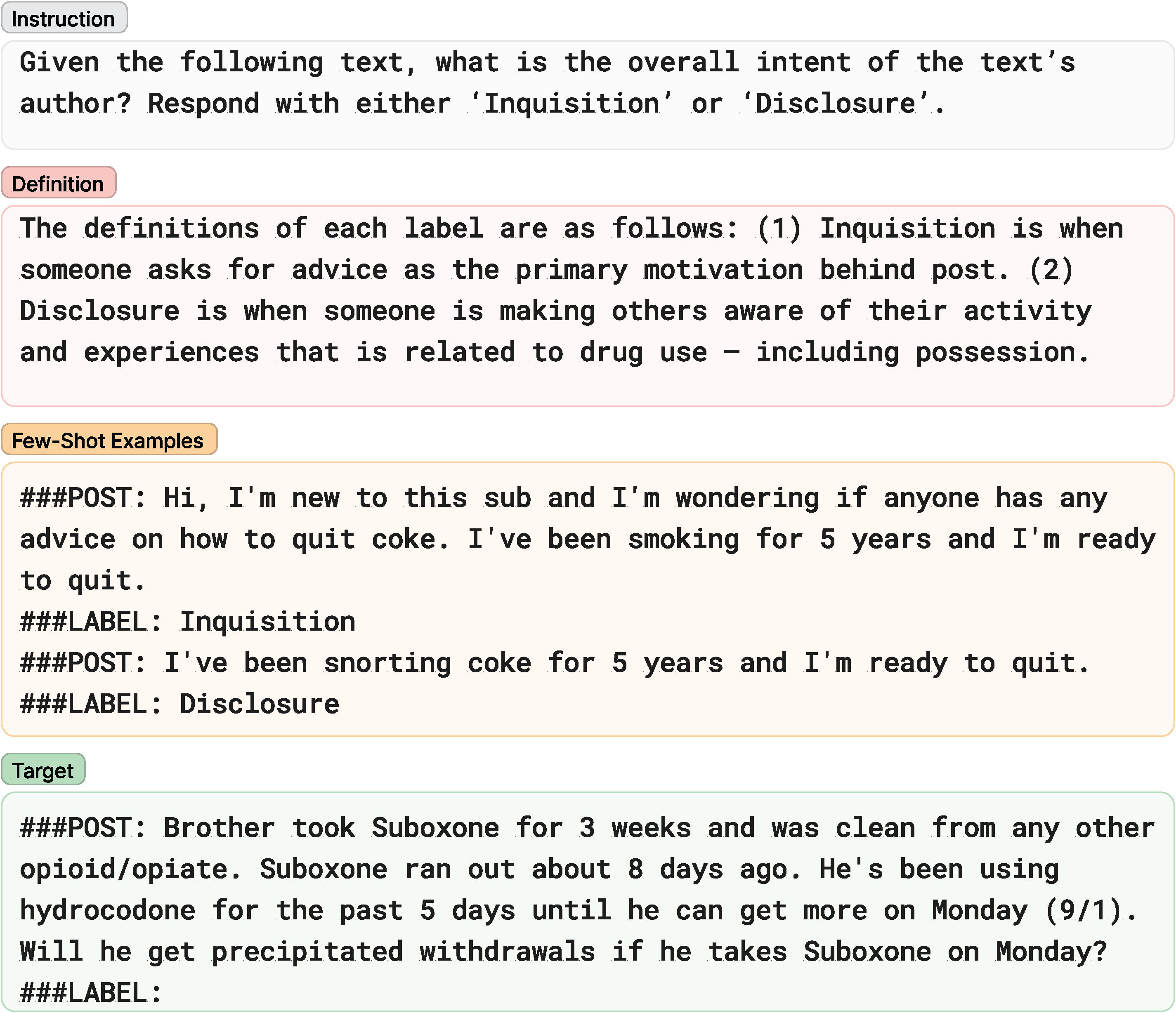}
    \caption{LLM prompting for Type of Connection}
    \label{fig:prompt}
\end{figure}

\vspace{-0.25cm}
\paragraph{LLM-based Models. }
To assess the effectiveness of LLMs in classifying texts within the domain of SUD, we conducted a comprehensive experiment involving three models: two proprietary models from OpenAI, GPT-3.5 (gpt-3.5-turbo-0613) and GPT-4 (gpt-4-0125-preview) \cite{openai2023gpt4}, and one open-source model, Mixtral 7B \cite{jiang2024mixtral}. We evaluated their performance across three prompting styles:
\begin{itemize}[noitemsep,topsep=0pt]
    \item[-] \textbf{Instruction-only (`I'):} This approach provided only the classification task instructions to the LLM.
    \item[-] \textbf{Instruction \& Definition (`I + D'): } Instructions with clear definitions of each classification label.
    \item[-] \textbf{Instruction, Definition, \& Examples (`I + D + E'): } Supplementing the definitions with two relevant examples for each label, mimicking few-shot learning scenarios.
\end{itemize}
Figure \ref{fig:prompt} illustrates the style of prompting used to label each post in our dataset. The dataset and models are available in our GitHub repository.\footnote{\url{https://github.com/social-nlp-lab/Drug-experience-classification}}

\section{Experiments Results}

% \usepackage{graphicx}
\begin{table*}[h]
% \small
\centering
\resizebox{\textwidth}{!}{%
\begin{tabular}{@{}cccccccccccccccccc@{}}\toprule
&  & \multicolumn{2}{c}{\textbf{\texttt{Connection}}} & \multicolumn{3}{c}{\textbf{\texttt{Subject}}} & \multicolumn{11}{c}{\textbf{\texttt{Objectives}}}\\ 
 & \multicolumn{1}{l}{\texttt{WC}} & \texttt{Inquisition} & \texttt{Disclosure} & \texttt{Dependency} & \texttt{Recovery} & \texttt{Other} & \texttt{C.o.S} & \texttt{E} & \texttt{L} & \texttt{M.H} & \texttt{M.o.I} & \texttt{N.S\&M} & \texttt{O} & \texttt{Q} & \texttt{R} & \texttt{S} & \texttt{W} \\\cmidrule(lr){2-2}\cmidrule(lr){3-4} \cmidrule(lr){5-7} \cmidrule(lr){8-18}  
\multicolumn{1}{l}{\texttt{Testing}}  & 106.1 & 51& 49 & 37 & 7  & 53  & 20 & 66  & 5  & 24 & 66  & 11 & 5 & 9  & 7  & 26 & 13 \\ \cmidrule(lr){1-1}
\multicolumn{1}{l}{\texttt{Training}} & 314.5 & 280& 118& 116& 7  & 256   & 77 & 185 & 12 & 38 & 199 & 37 & 4 & 32 & 14 & 55 & 37 \\ \bottomrule
\multicolumn{1}{l}{\texttt{Total}} & 147.7 & 331& 167& 153& 14  & 309   & 97 & 251 & 17 & 62 & 265 & 48 & 9 & 41 & 21 & 81 & 50 \\ \bottomrule
\end{tabular}%
}
\caption{Frequencies of classes in each level labeled within training and test sets. Abbreviations: WC: word count; C.o.S: combination of substances; E: effects; L: legality; M.H.: mental health; M.o.I: methods of ingestion; N.S\&M: nurturant support \& morality; O: overdose; Q: quality; R: relapse; S: safety; W: withdrawal. }
\label{tab:freq}
\end{table*}

\begin{table*}[h!]
\centering
\resizebox{\linewidth}{!}{%
\begin{tabular}{@{}lllllllllllll@{}}
\toprule
\multicolumn{3}{c}{\texttt{\textbf{}}} & \multicolumn{3}{c}{\texttt{\textbf{Connection}}} & \multicolumn{3}{c}{\texttt{\textbf{Subject}}} & \multicolumn{3}{c}{\texttt{\textbf{Objective}}}\\ 
 \cmidrule(lr){4-6} \cmidrule(lr){7-9} \cmidrule(lr){10-12}
\texttt{\textbf{}} & \texttt{\textbf{Feature}} & \texttt{\textbf{Classifiers}} & \texttt{\textbf{Prec}} & \texttt{\textbf{Rec}} & \texttt{\textbf{F1}} & \texttt{\textbf{Prec}} & \texttt{\textbf{Rec}} & \texttt{\textbf{F1}} & \texttt{\textbf{Prec}} & \texttt{\textbf{Rec}} & \texttt{\textbf{F1}}\\ 
\cmidrule(lr){1-3}\cmidrule(lr){4-6} \cmidrule(l){7-9} \cmidrule(l){10-12}

\multirow{4}{*}{\texttt{Baseline}} & \texttt{TF-IDF} & \texttt{KNN} & 0.77 & 0.72 & 0.68 & 0.49 & 0.51 & 0.49 & 0.35 & 0.44 & 0.39\\
 & \texttt{BERT} & \texttt{LogR} & 0.71 & 0.72 & 0.71 & 0.53 & 0.58 & 0.55 & 0.60 & 0.47 & 0.49\\
 & \texttt{BERT} & \texttt{SVM} & 0.75 & 0.75 & 0.75 & 0.56 & 0.60 & 0.58 & 0.52 & 0.45 & 0.44\\ 
 & \texttt{ada-002} & \texttt{LogR} & 0.82 & 0.82 & 0.81 & 0.57 & 0.63 & 0.58 & 0.48 & 0.30 & 0.34\\
 \cmidrule(lr){1-3}
\multirow{2}{*}{\texttt{Transformer-Based}} & \texttt{DeBERTa} & \texttt{DeBERTa} & \textbf{0.95} & 0.87 & 0.90 & 0.66 & 0.71 & 0.69 & 0.51 & 0.44 & 0.43\\
 & \texttt{RoBERTa} & \texttt{RoBERTa} & 0.85 & 0.81 & 0.83 & 0.63 & 0.69 & 0.66 & 0.39 & 0.43 & 0.41\\ 

\texttt{} & - & \texttt{SetFit} & 0.90 & 0.64 & 0.75 & 0.62 & 0.39 & 0.45 & 0.37 & 0.43 & 0.39\\ 
\cmidrule(lr){1-3}
\multirow{2}{*}{\texttt{LLM}} & - & \texttt{GPT4 I + D} & 0.76 & 0.45 & 0.32 & \textbf{0.77} & 0.64 & 0.66 & 0.56 & \textbf{0.76} & 0.61 \\
 & - & \texttt{GPT4 I + D + E} & 0.91 & \textbf{0.91} & \textbf{0.91} & 0.74 & \textbf{0.72} & \textbf{0.73} & \textbf{0.71} & 0.66 & \textbf{0.64}\\
 & - & \texttt{Mixtral I + D + E} & 0.78 & 0.51 & 0.43 & 0.72 & 0.34 & 0.23 & 0.42 & 0.84 & 0.53\\
\bottomrule
\end{tabular}%
}
\caption{Weighted Precision, Recall, F1 scores for each level of experience for performant classification models. The best score for each metric is in \textbf{bold}. Metrics for all experiments can be found in Appendix \ref{tab:full-results}.}
\label{tab:results}
\end{table*}

\subsection{Data Analysis}
We present the distribution of classes in Table~\ref{tab:freq}, which was derived by applying our taxonomy to 500 posts. Between the testing and training sets, approximately 66.2\% (N = 331) of the type of Connection were labeled as Inquisition, and 61\% were labeled as Other for Subject. Among the Objectives, Methods of Ingestion (N = 265, 53\%) and Effects (N = 251, 50.1\%) were the most prevalent, while Overdose (N = 9, 1.8\%) was the least prevalent.

\subsection{Classification}
Table \ref{tab:results} summarizes a selected set of results of our experiments using (1) baseline models, (2) Transformer-based models, and (3) LLMs-based models. As shown in the table, for the binary task of Connection (Inclusion vs. Disclosure), DeBERTa outperformed all models with respect to precision at 0.95, indicating its effectiveness in returning more relevant results overall.  
Using a Few-Shot model did not increase the performance compared to the baseline. However, the GPT-4 model with `I + D + E' achieved the highest Recall (0.91) and F1 scores (0.91) among all models.

We observed a similar trend in the Subject level, which includes more fine-grained labels (i.e., Dependency, Recovery, Other, and N/A). Specifically, the GPT-4 model using the `I + D + E' method achieved the highest recall (0.72) and F1 score (0.73), while the GPT-4 model with the `I + D' strategy outperformed the other models in terms of precision (0.77), with around 5\% improvement.

Similarly, for the Objective classes, the best performing model was GPT4 with `I + D + E' with respect to precision (0.71) and F1 (0.64). However, the GPT4 model with `I + D' outperformed other models in recall (0.76), suggesting its ability to correctly identify a higher proportion of relevant instances. 
A more detailed discussion of the classification procedure and results is provided in Appendix \ref{app:full_result} and Table \ref{tab:full-results}.

% \subsection{Linguistic Analysis}

% In Table X, 
\section{Usefulness of our Models}
In this study, we introduce a new taxonomy to better understand the lived experiences of PWUD online and analyze these experiences in user-generated texts on Reddit.  
Our study builds upon existing literature that employed qualitative analysis to uncover elements of social support within drug-focused Reddit communities \cite{bunting_socially-supportive_2021,gauthier_i_2022, dagostino_social_2017,graves_thematic_2022, wombacher_social_2020}. However, we apply computational methods to conduct this exploration, a novel approach compared to previous studies that have primarily concentrated on binary classifications \cite{al-garadi_text_2021} or other aspects of mental health \cite{balani_detecting_2015,gaur_let_2018,valdez_computational_2022}. Our experiments demonstrate significant performance differences between classical machine learning, state-of-the-art transformer-based models, and LLMs for multi-label classification of drug-related posts. LLMs outperformed other models in detecting diverse aspects of lived experience disclosures, underscoring their potential for larger-scale investigations into personal narratives of drug use.

Current U.S. addiction recovery frameworks often fail to consider the lived experiences of individuals with SUD \cite{samhsa2016}, assuming universal access to professional guidance—a reality not shared by many affected. Online communities fill this gap, offering acceptance, understanding, and validation not always present in professional settings, thus becoming crucial for those lacking healthcare access \cite{mead2006peer}. They enable discussions on substance use and recovery, playing a vital role for individuals isolated from traditional healthcare resources\cite{Boisvert_Martin_Grosek_Clarie_2008}. Our study aims to address these knowledge gaps by detailing the experiences of PWUD, enhancing the ability of experts to offer more comprehensive and personalized support.
Our results show a balanced mix of users seeking (Inquisition) and sharing (Disclosure) information, aligning with prior work \cite{valdez_computational_2022}. Analyzing self-disclosed experiences within the posts indicates a minor portion discussing recovery experiences, touching on Nurturant Support, Relapse, and Safety. This finding is expected as recovery subreddits like \textit{r/OpiatesRecovery}, and \textit{r/benzorecovery}, host more detailed recovery discussions. 

Prior research identified nurturant support themes of recovery and addiction in the \textit{r/Drugs} subreddit\cite{wombacher_social_2020}. Our study builds on this, offering deeper insights into the nuanced discussions of recovery and dependency, as shown in this post:

\textit{``171 days sober, I last posted 140 days ago about how getting cleaned up from opiates has revolutionized my life. just wanted to check in with everyone and see how everyone's doing.''}

Among posts that disclosed a Dependency experience (N = 153), most contained themes of Effects and Methods of Ingestion, resonating with the concept of action-facilitating support identified in the \textit{r/Drugs} community \cite{wombacher_social_2020}: 

\textit{``... I was on opiates. I started at ten-mg hydro twice a day for 2-3 months, after being directed to a pain doctor they were able to move me to 10mg/325 Percocet 4x a day since July...With all the horrible stigma surrounding opiates, I'm anxious to talk to my doctor about bumping or changing my meds...My tolerance has gone...I sometimes notice WD symptoms when I wait too long''}

This relationship confirms the value of online forums as spaces for sharing and obtaining information on substance use and recovery, highlighting our study's relevance. Moreover, examining conversations about Ingestion Methods and Effects can aid harm reduction. Sharing insights on safer practices, dosages, and possible side effects can help users reduce substance use risks.

\vspace{-0.3cm}
\subsection{Error Analysis}
To gain a more comprehensive understanding of the best model's performance, we conducted a detailed error analysis. Despite the notable performance of GPT-4 `I + D + E' in classifying a broad range of classes across all three levels, we found some misclassified instances in the test set. For posts tagged as Connection, out of 22 tagged as Disclosure, only 3 were incorrectly classified as Inquisition, and from the 31 posts tagged as Inquisition, 2 were misclassified as Disclosure. At the Subject level, of the 17 posts tagged as Dependency, 4 were incorrectly classified as Recovery. Examples of these misclassified instances are presented in Table \ref{tab:label_comparison}.

Our analysis shows that the model sometimes had difficulty discerning the subtle intentions expressed in personal narratives about drug experiences. This was particularly true in instances where narratives combined Disclosure (sharing of personal substance use experiences) with a question for feedback or solutions, not directly related to SUD. The dual nature of these communications often made it challenging for the model to determine the dominant intent behind the posts.
Further analysis at the Subject level identified cases where the model incorrectly classified posts revolved around Dependency as Recovery. In some instances, even brief mentions of sobriety within an individual's narrative were interpreted by the model as signs of Recovery. This underscores the model's tendency to occasionally misjudge the context or importance of specific keywords in the discourse.
Furthermore, our findings indicate a discrepancy in the Objective level classification, stemming from the model's inclination to over-rely on specific keywords while neglecting the wider context, potentially skewing the narratives. 
This skewed portrayal of user narratives underscores the need for domain experts to be involved in promoting and evaluating models, thereby enhancing the model's ability to understand and interpret context more effectively.

\section{The Language of Personal Drug Experiences}

\begin{table*}[htbp]
% \centering
% \resizebox{0.6\textwidth}{!}{%
\centering
\resizebox{0.65\textwidth}{!}{%
\begin{tabular}{@{}lccccccccccc|c@{}}
\toprule
 & \multicolumn{11}{c|}{\texttt{$\mu_{Inquisition}/\mu_{Disclosure}$}} & \texttt{$\mu_{Recovery}/\mu_{Dependency}$} \\ \cmidrule(lr){2-12} \cmidrule(lr){13-13}
\rotatebox{90}{\texttt{LWC}} & \rotatebox{90}{\texttt{Recovery}} & \rotatebox{90}{\texttt{Dependency}} & \rotatebox{90}{\texttt{Effects}} & \rotatebox{90}{\texttt{Methods of Ingestion}} & \rotatebox{90}{\texttt{Comb. of Substances}} & \rotatebox{90}{\texttt{Mental Health}} & \rotatebox{90}{\texttt{Nurturant \& Morality}} & \rotatebox{90}{\texttt{Withdrawal}} & \rotatebox{90}{\texttt{Safety}} & \rotatebox{90}{\texttt{Relapse}} & \rotatebox{90}{\texttt{ALL}} &\rotatebox{90}{\texttt{Effects}} \\ \midrule
\textbf{Qmark} & \cellcolor[HTML]{D65C5E}6.39 & \cellcolor[HTML]{D65C5E}5.19 & \cellcolor[HTML]{F5DCDC}1.15 & \cellcolor[HTML]{E6B5B6}1.28 & \cellcolor[HTML]{DDEBF7}0.85 & \cellcolor[HTML]{DDEBF7}0.94 & \cellcolor[HTML]{D65C5E}2.79 & \cellcolor[HTML]{D65C5E}5.03 & \cellcolor[HTML]{DDEBF7}0.86 & \cellcolor[HTML]{D65C5E}3.48 & \cellcolor[HTML]{D65C5E}2.70 & \cellcolor[HTML]{FFFFFF} \\
\textbf{illness} & \cellcolor[HTML]{FFFFFF} & \cellcolor[HTML]{FFFFFF} & \cellcolor[HTML]{D65C5E}3.03 & \cellcolor[HTML]{FFFFFF} & \cellcolor[HTML]{D65C5E}3.03 & \cellcolor[HTML]{FFFFFF} & \cellcolor[HTML]{FFFFFF} & \cellcolor[HTML]{FFFFFF} & \cellcolor[HTML]{D65C5E}2.97 & \cellcolor[HTML]{FFFFFF} & \cellcolor[HTML]{FFFFFF} & \cellcolor[HTML]{FFFFFF} \\
\textbf{feeling} & \cellcolor[HTML]{FFFFFF} & \cellcolor[HTML]{FFFFFF} & \cellcolor[HTML]{FFFFFF} & \cellcolor[HTML]{FFFFFF} & \cellcolor[HTML]{D65C5E}2.56 & \cellcolor[HTML]{FFFFFF} & \cellcolor[HTML]{FFFFFF} & \cellcolor[HTML]{FFFFFF} & \cellcolor[HTML]{FFFFFF} & \cellcolor[HTML]{FFFFFF} & \cellcolor[HTML]{FFFFFF} & \cellcolor[HTML]{FFFFFF} \\
\textbf{risk} & \cellcolor[HTML]{FFFFFF} & \cellcolor[HTML]{FFFFFF} & \cellcolor[HTML]{FFFFFF} & \cellcolor[HTML]{FFFFFF} & \cellcolor[HTML]{D65C5E}1.98 & \cellcolor[HTML]{FFFFFF} & \cellcolor[HTML]{FFFFFF} & \cellcolor[HTML]{FFFFFF} & \cellcolor[HTML]{D65C5E}1.98 & \cellcolor[HTML]{FFFFFF} & \cellcolor[HTML]{FFFFFF} & \cellcolor[HTML]{FFFFFF} \\
\textbf{tentat} & \cellcolor[HTML]{FFFFFF} & \cellcolor[HTML]{D65C5E}1.73 & \cellcolor[HTML]{D65C5E}1.97 & \cellcolor[HTML]{D65C5E}2.14 & \cellcolor[HTML]{D65C5E}2.24 & \cellcolor[HTML]{E6B5B6}1.44 & \cellcolor[HTML]{FFFFFF} & \cellcolor[HTML]{FFFFFF} & \cellcolor[HTML]{D65C5E}2.21 & \cellcolor[HTML]{FFFFFF} & \cellcolor[HTML]{D65C5E}2.08 & \cellcolor[HTML]{FFFFFF} \\
\textbf{curiosity} & \cellcolor[HTML]{FFFFFF} & \cellcolor[HTML]{FFFFFF} & \cellcolor[HTML]{FFFFFF} & \cellcolor[HTML]{E6B5B6}1.33 & \cellcolor[HTML]{FFFFFF} & \cellcolor[HTML]{FFFFFF} & \cellcolor[HTML]{FFFFFF} & \cellcolor[HTML]{FFFFFF} & \cellcolor[HTML]{E6B5B6}1.39 & \cellcolor[HTML]{FFFFFF} & \cellcolor[HTML]{D65C5E}2.02 & \cellcolor[HTML]{FFFFFF} \\
\textbf{discrep} & \cellcolor[HTML]{FFFFFF} & \cellcolor[HTML]{E6B5B6}1.39 & \cellcolor[HTML]{E6B5B6}1.27 & \cellcolor[HTML]{D65C5E}1.62 & \cellcolor[HTML]{D65C5E}1.87 & \cellcolor[HTML]{D65C5E}1.74 & \cellcolor[HTML]{FFFFFF} & \cellcolor[HTML]{FFFFFF} & \cellcolor[HTML]{E6B5B6}1.52 & \cellcolor[HTML]{FFFFFF} & \cellcolor[HTML]{D65C5E}1.59 & \cellcolor[HTML]{FFFFFF} \\
\textbf{polite} & \cellcolor[HTML]{FFFFFF} & \cellcolor[HTML]{D65C5E}2.39 & \cellcolor[HTML]{D65C5E}2.23 & \cellcolor[HTML]{D65C5E}2.70 & \cellcolor[HTML]{D65C5E}7.61 & \cellcolor[HTML]{D65C5E}5.80 & \cellcolor[HTML]{FFFFFF} & \cellcolor[HTML]{FFFFFF} & \cellcolor[HTML]{D65C5E}4.75 & \cellcolor[HTML]{FFFFFF} & \cellcolor[HTML]{E6B5B6}1.51 & \cellcolor[HTML]{D65C5E}1.72 \\
\textbf{cause} & \cellcolor[HTML]{FFFFFF} & \cellcolor[HTML]{D65C5E}1.82 & \cellcolor[HTML]{E6B5B6}1.40 & \cellcolor[HTML]{D65C5E}1.54 & \cellcolor[HTML]{D65C5E}1.73 & \cellcolor[HTML]{F5DCDC}1.20 & \cellcolor[HTML]{FFFFFF} & \cellcolor[HTML]{FFFFFF} & \cellcolor[HTML]{E6B5B6}1.47 & \cellcolor[HTML]{FFFFFF} & \cellcolor[HTML]{E6B5B6}1.51 & \cellcolor[HTML]{FFFFFF} \\
\textbf{insight} & \cellcolor[HTML]{FFFFFF} & \cellcolor[HTML]{E6B5B6}1.42 & \cellcolor[HTML]{F5DCDC}1.12 & \cellcolor[HTML]{F5DCDC}1.21 & \cellcolor[HTML]{DDEBF7}0.93 & \cellcolor[HTML]{F5DCDC}1.25 & \cellcolor[HTML]{FFFFFF} & \cellcolor[HTML]{FFFFFF} & \cellcolor[HTML]{DDEBF7}0.99 & \cellcolor[HTML]{FFFFFF} & \cellcolor[HTML]{E6B5B6}1.48 & \cellcolor[HTML]{FFFFFF} \\
\textbf{emo\_sad} & \cellcolor[HTML]{FFFFFF} & \cellcolor[HTML]{FFFFFF} & \cellcolor[HTML]{FFFFFF} & \cellcolor[HTML]{FFFFFF} & \cellcolor[HTML]{FFFFFF} & \cellcolor[HTML]{FFFFFF} & \cellcolor[HTML]{FFFFFF} & \cellcolor[HTML]{FFFFFF} & \cellcolor[HTML]{FFFFFF} & \cellcolor[HTML]{FFFFFF} & \cellcolor[HTML]{DDEBF7}0.86 & \cellcolor[HTML]{D65C5E}3.43 \\
\textbf{tone\_pos} & \cellcolor[HTML]{FFFFFF} & \cellcolor[HTML]{FFFFFF} & \cellcolor[HTML]{FFFFFF} & \cellcolor[HTML]{FFFFFF} & \cellcolor[HTML]{FFFFFF} & \cellcolor[HTML]{FFFFFF} & \cellcolor[HTML]{FFFFFF} & \cellcolor[HTML]{FFFFFF} & \cellcolor[HTML]{FFFFFF} & \cellcolor[HTML]{FFFFFF} & \cellcolor[HTML]{5B9BD5}0.58 & \cellcolor[HTML]{FFFFFF} \\
\textbf{money} & \cellcolor[HTML]{FFFFFF} & \cellcolor[HTML]{FFFFFF} & \cellcolor[HTML]{FFFFFF} & \cellcolor[HTML]{FFFFFF} & \cellcolor[HTML]{FFFFFF} & \cellcolor[HTML]{FFFFFF} & \cellcolor[HTML]{FFFFFF} & \cellcolor[HTML]{FFFFFF} & \cellcolor[HTML]{FFFFFF} & \cellcolor[HTML]{FFFFFF} & \cellcolor[HTML]{5B9BD5}0.57 & \cellcolor[HTML]{FFFFFF} \\
\textbf{home} & \cellcolor[HTML]{FFFFFF} & \cellcolor[HTML]{FFFFFF} & \cellcolor[HTML]{5B9BD5}0.23 & \cellcolor[HTML]{5B9BD5}0.25 & \cellcolor[HTML]{5B9BD5}0.09 & \cellcolor[HTML]{FFFFFF} & \cellcolor[HTML]{FFFFFF} & \cellcolor[HTML]{5B9BD5}0.12 & \cellcolor[HTML]{FFFFFF} & \cellcolor[HTML]{FFFFFF} & \cellcolor[HTML]{5B9BD5}0.50 & \cellcolor[HTML]{D65C5E}2.92 \\
\textbf{we} & \cellcolor[HTML]{FFFFFF} & \cellcolor[HTML]{FFFFFF} & \cellcolor[HTML]{5B9BD5}0.37 & \cellcolor[HTML]{FFFFFF} & \cellcolor[HTML]{FFFFFF} & \cellcolor[HTML]{FFFFFF} & \cellcolor[HTML]{FFFFFF} & \cellcolor[HTML]{5B9BD5}0.09 & \cellcolor[HTML]{5B9BD5}0.09 & \cellcolor[HTML]{FFFFFF} & \cellcolor[HTML]{5B9BD5}0.49 & \cellcolor[HTML]{FFFFFF} \\
\textbf{family} & \cellcolor[HTML]{FFFFFF} & \cellcolor[HTML]{DDEBF7}0.72 & \cellcolor[HTML]{5B9BD5}0.60 & \cellcolor[HTML]{5B9BD5}0.60 & \cellcolor[HTML]{FFFFFF} & \cellcolor[HTML]{FFFFFF} & \cellcolor[HTML]{FFFFFF} & \cellcolor[HTML]{5B9BD5}0.57 & \cellcolor[HTML]{FFFFFF} & \cellcolor[HTML]{FFFFFF} & \cellcolor[HTML]{5B9BD5}0.47 & \cellcolor[HTML]{E6B5B6}1.49 \\
\textbf{affiliation} & \cellcolor[HTML]{FFFFFF} & \cellcolor[HTML]{FFFFFF} & \cellcolor[HTML]{FFFFFF} & \cellcolor[HTML]{FFFFFF} & \cellcolor[HTML]{FFFFFF} & \cellcolor[HTML]{FFFFFF} & \cellcolor[HTML]{FFFFFF} & \cellcolor[HTML]{5B9BD5}0.45 & \cellcolor[HTML]{FFFFFF} & \cellcolor[HTML]{FFFFFF} & \cellcolor[HTML]{FFFFFF} & \cellcolor[HTML]{FFFFFF} \\
\textbf{moral} & \cellcolor[HTML]{FFFFFF} & \cellcolor[HTML]{FFFFFF} & \cellcolor[HTML]{FFFFFF} & \cellcolor[HTML]{FFFFFF} & \cellcolor[HTML]{FFFFFF} & \cellcolor[HTML]{FFFFFF} & \cellcolor[HTML]{FFFFFF} & \cellcolor[HTML]{FFFFFF} & \cellcolor[HTML]{FFFFFF} & \cellcolor[HTML]{FFFFFF} & \cellcolor[HTML]{5B9BD5}0.42 & \cellcolor[HTML]{FFFFFF} \\
\textbf{emo\_pos} & \cellcolor[HTML]{FFFFFF} & \cellcolor[HTML]{FFFFFF} & \cellcolor[HTML]{5B9BD5}0.38 & \cellcolor[HTML]{FFFFFF} & \cellcolor[HTML]{FFFFFF} & \cellcolor[HTML]{FFFFFF} & \cellcolor[HTML]{FFFFFF} & \cellcolor[HTML]{FFFFFF} & \cellcolor[HTML]{FFFFFF} & \cellcolor[HTML]{FFFFFF} & \cellcolor[HTML]{5B9BD5}0.41 & \cellcolor[HTML]{FFFFFF} \\
\textbf{memory} & \cellcolor[HTML]{FFFFFF} & \cellcolor[HTML]{FFFFFF} & \cellcolor[HTML]{5B9BD5}0.38 & \cellcolor[HTML]{FFFFFF} & \cellcolor[HTML]{FFFFFF} & \cellcolor[HTML]{FFFFFF} & \cellcolor[HTML]{FFFFFF} & \cellcolor[HTML]{FFFFFF} & \cellcolor[HTML]{FFFFFF} & \cellcolor[HTML]{FFFFFF} & \cellcolor[HTML]{5B9BD5}0.40 & \cellcolor[HTML]{FFFFFF} \\
\textbf{reward} & \cellcolor[HTML]{FFFFFF} & \cellcolor[HTML]{FFFFFF} & \cellcolor[HTML]{FFFFFF} & \cellcolor[HTML]{FFFFFF} & \cellcolor[HTML]{FFFFFF} & \cellcolor[HTML]{FFFFFF} & \cellcolor[HTML]{FFFFFF} & \cellcolor[HTML]{FFFFFF} & \cellcolor[HTML]{FFFFFF} & \cellcolor[HTML]{FFFFFF} & \cellcolor[HTML]{5B9BD5}0.39 & \cellcolor[HTML]{D65C5E}5.71 \\
\textbf{wellness} & \cellcolor[HTML]{FFFFFF} & \cellcolor[HTML]{FFFFFF} & \cellcolor[HTML]{FFFFFF} & \cellcolor[HTML]{FFFFFF} & \cellcolor[HTML]{FFFFFF} & \cellcolor[HTML]{FFFFFF} & \cellcolor[HTML]{FFFFFF} & \cellcolor[HTML]{FFFFFF} & \cellcolor[HTML]{FFFFFF} & \cellcolor[HTML]{FFFFFF} & \cellcolor[HTML]{5B9BD5}0.30 & \cellcolor[HTML]{D65C5E}3.74
\\
\bottomrule
\end{tabular}%
}
\caption{Top 20 largest and smallest LIWC categories by effect size. Highlighted cells indicate different ratios: \colorbox[HTML]{5B9BD5}{dark blue} represents the lowest ratios (smaller effect sizes), and \colorbox[HTML]{D65C5E}{dark red} represents the highest ratios (larger effect sizes). Empty cells denote non-significant results.}
\label{tab:liwc1}
% \end{minipage} 
\vspace{-0.5 cm}
\end{table*}
We applied our most effective model, GPT4 `I + D + E', to extend our analysis to 1,000 randomly selected posts from our dataset for psycholinguistic analysis of PWUD's lived experiences. To ensure a more representative sample and increase statistical power, we augmented the pre-annotated set of 500 posts with 1,000 randomly selected posts. This resulted in a comprehensive analysis of 1,500 posts across three dimensions: Connection, Subject, and Objective. Table \ref{tab:cross-tab} presents the distribution of labels across our dataset. We applied the Linguistic Inquiry and Word Count (LIWC) tool \cite{boyd2022development}, which quantifies the prevalence of words from diverse categories in the text, to this dataset. To compare the average scores for LIWC categories, we conducted non-parametric Mann-Whitney U tests \cite{Mann_Whitney_1947}, identifying statistically significant differences (with $p < .05$) between Inquisition vs. Disclosure and Recovery vs. Dependency posts. The Benjamini-Hochberg procedure \cite{benjamini} was employed to control the false discovery rate across 85 selected LIWC categories. Our analysis of posts considered two metrics: (1) the ratio for posts labeled as Recovery and Dependency ($R_R = \mu_{Recovery}/\mu_{Dependency}$), and (2) the ratio of mean LIWC category scores between Inquisition and Disclosure posts ( $R_I = \mu_{Inquisition}/\mu_{Disclosure}$). Detailed results can be found in Appendix \ref{app:LIWC}. We found that 47 LIWC categories were significantly associated with Inquisition posts ($p < .05$) overall, while six were significantly associated with Recovery posts. Table \ref{tab:liwc1} presents the LIWC categories with the largest and smallest effect sizes (i.e., the ratios farthest from 1).

\vspace{-0.2cm}
\paragraph{Inquisition vs. Disclosure. } 

As expected, posts seeking information (Inquisition) used significantly more question marks (\textit{Qmark}). However, this pattern was less pronounced in posts discussing Safety (0.94), Combination of Substances (0.84), and Mental Health (0.86), suggesting users disclose more information about these topics. Notably, we observed 1.16 times greater \textit{authenticity} in Inquisition posts compared to Disclosure posts. 
Posts discussing Withdrawal experiences were less inquisitive, indicating a preference for sharing personal experiences rather than seeking information on this sensitive topic. 
Higher usage of LIWC categories such as \textit{friend}, \textit{conflict}, \textit{affiliation} further support this interpretation (Table \ref{tab:LIWC}).

The use of \textit{prosocial} language (e.g., ``care'', ``help'') was 2.54 times more frequent in posts inquiring about Dependency concerns. This may suggest that individuals seeking information about dependency may frame their inquiries in a way that elicits support and empathy. We found \textit{prosocial} language also more prevalent in Disclosure posts related to Combination of Substances (0.46), indicating that discussions of polysubstance use may involve a greater degree of mutual support.
While the use of language to establish \textit{Clout} was generally more prevalent in Disclosure posts (0.89), likely reflecting a desire to establish credibility when sharing personal experiences, this pattern was particularly notable in discussions of Methods of Ingestion (0.69), suggesting that individuals sharing experiences related to drug use methods may be especially motivated to present themselves as knowledgeable and authoritative. 

Finally, while our analysis revealed significant differences in the use of \textit{relig} and \textit{food} categories, further investigation highlighted the limitations of LIWC limitations in capturing the nuances of drug-related discourse. These terms often appear in metaphorical contexts within drug discourse (e.g., ``god,'' ``hell,'' ``cook,'' ``bake''), underscoring the need for specialized linguistic tools tailored to the unique language of online drug forums.

\vspace{-0.2cm}
\paragraph{Combination of Substances and Safety. } 

Posts inquiring about combining substances used significantly more \textit{polite} language (7.6 times more) than those disclosing experiences. This suggests a strategic use of politeness to foster a supportive environment when seeking potentially sensitive information. %Interestingly, the heightened 
More frequent use of \textit{acquire} language (e.g., ``get,'' ``take'') in these inquiries could reflect a desire for information and a potential interest in consuming multiple substances. The threefold increase in \textit{illness} language within posts about physiological Effects, Combination of Substances, and Safety suggests that users are prioritizing harm reduction and seeking information about potential negative health consequences. The significant association of \textit{risk} language specifically with posts about combining substances and Safety underscores the growing concern users may have regarding the potential dangers of polysubstance use.

\vspace{-0.2cm}
\paragraph{Discussion of Physiological Effects. } 

The use of \textit{polite} language is approximately 30\% more frequent in Inquisition posts about physiological Effects (ratio of 2.23) compared to Recovery posts (ratio of 1.72). This difference may reflect a greater degree of deference or caution when seeking information about potentially stigmatized experiences. On the other hand, the more prevalent use of \textit{emo\_sad} language in Recovery posts about effects (3.4 times more than in Dependency posts) suggests that individuals sharing personal experiences related to recovery may be more likely to express negative emotions associated with the physiological consequences of drug use. The increased use of \textit{home} (e.g., ``home,'' ``bed'') and \textit{family} (e.g., ``mother,'' ``brother'') language in Recovery posts, which were 2.9 and 1.49 times more frequent respectively, suggests that discussions of recovery often involve reflections on personal relationships and living environments, potentially highlighting the importance of social support and stable environments in the recovery process. The higher use of \textit{reward} and \textit{wellness} language in Recovery posts compared to Dependency posts suggests a focus on positive outcomes and potential benefits of overcoming physiological dependencies.

\section{Conclusion}

This study aims to identify how individuals discuss their personal drug experiences online. Using a deductive-inductive approach, we developed a taxonomy to assess user-generated posts, including their intended Connections (Inquisition or Disclosure), Subjects (e.g., Recovery, Dependency, Other), and specific Objectives (e.g., Relapse, Quality, Safety, Legality). We then employed this taxonomy to annotate 500 randomly sampled posts from a dataset we created, consisting of posts from four subreddits: \textit{r/opiates}, \textit{r/benzodiazepines}, \textit{r/stims}, and \textit{r/cocaine}. We used this data to train three sets of classifiers: (1) baseline models, (2) transformer-based deep learning models, and (3) LLM-based models including GPT-3.5 Turbo, GPT-4, and Mixtral, an open-source LLM. Our data analysis shows that posts are more inquisitive and predominately revolve around Effects and Methods of Ingestion. Our classification results show that GPT-4 with Instruction, Definition, and Examples (`I +D+ E') in prompts outperformed other models, whereas the DeBERTa-based transformer model had the best performance among the non-LLM models. 

After applying our best-performing model to an additional 1,000 randomly selected posts, we used LIWC to analyze the linguistic differences between posts labeled as Inquisition vs. Disclosure and Recovery vs. Dependency. This analysis revealed that Inquisition posts significantly used more \textit{authenticity}-related language, while Disclosure posts emphasized personal sharing, especially in sensitive topics like Withdrawal. The analysis also highlighted the varied use of \textit{prosocial}, \textit{clout}, and metaphorical language across different discussion themes, providing deeper insights into the psychosocial dynamics of online drug-related discourse.

These findings provide insight into the intricate language of drug use discussions, highlighting potential indicators of SUD or recovery initiation. Our results underscore how online forums provide crucial support and safety planning, demonstrating the value of computational analysis in understanding health-related online communities and informing SUD treatment and recovery interventions.

\section*{Acknowledgements}
We would like to thank Dr. Robert Sterling, Lauren Kairys, and Maggie Dickinson for their assistance with evaluating our codebook and their insightful feedback. We thank Sanonda Gupta for her help and Satvik Bahsin, Amui Gayle, Donald Hattier, Lauren Miller, Linh Nguyen, and Medhavi Pandit for all their hard work in the development of the codebook used for annotation. We also want to acknowledge the Reddit users whose discussions on various aspects of drug use served as the backbone of this study.

\section{Limitations}
Thematic analysis, inherently subjective, depends on the annotator's interpretation, potentially introducing biases into the coding process and affecting data accuracy. Focusing solely on four subreddits to derive insights into the SUD and recovery communities brings limitations, given the existence of numerous forums on these topics. Our selection aimed to reflect the broader nuances within the SUD landscape, acknowledging the constraints of such a scope.

Human annotation, while detailed, poses challenges due to its time-intensive nature for large datasets, limiting the data volume for model training and possibly affecting the models' performance and applicability. Moreover, the efficacy of few-shot learning is contingent upon the quality and diversity of training data; limitations include the risk of overfitting, computational inefficiencies, and poor generalization to new tasks or data distributions in cases of noisy or insufficient data.

Annotator unfamiliarity with specific substances could lead to misinterpretations that experts might avoid, underscoring the models' goal to grasp context without always requiring domain expertise. Additionally, the predominantly pseudonymous nature of Reddit participation, skewing towards a younger, majority male demographic, coupled with \newcite{Hargittai_2020} observation that social media users tend to be of higher socioeconomic status, suggests a potential skew in the perspectives represented in computational social science research.

Our classification models excel in different levels of categorization but struggle with fine-grained, domain-specific classes, leading to potential misclassifications or oversimplifications. This limitation poses a challenge to the accurate recognition of nuanced distinctions within posts. Enhancing the models' capabilities in identifying these detailed classes (e.g., by using external knowledge bases) is essential for improving the utility of our models in domain-specific applications, thereby enabling more precise analysis and targeted interventions. 
Future work will focus on addressing these limitations. 

\section{Ethics Statement}

In alignment with the harm reduction perspective, which prioritizes the autonomy and well-being of PWUD \cite{harmreductioncoalition2024}, our research methodology places significant importance on protecting the privacy and anonymity of Reddit users who share their personal experiences. This study has received approval from the University's Institutional Review Board (IRB), ensuring adherence to rigorous ethical standards. To further safeguard user privacy, we will not release the full text of Reddit posts. Instead, we will only make available post IDs alongside their corresponding labels, allowing other researchers with appropriate access to reproduce our findings while ensuring user anonymity. Additionally, any quotes used in this work have been carefully modified to remove identifying details and protect user privacy. This involved paraphrasing content, removing specific drug names or slang terms, and generalizing language used in the quotes. We believe this approach upholds the principles of harm reduction while enabling valuable research to be conducted in a manner that respects the dignity and confidentiality of the individuals who generously share their experiences.

\bibliography{custom,references}
\bibliographystyle{acl_natbib}

\appendix

\section{Taxonomy of Personal Drug Experiences}
\label{appendix:definitions}
As discussed in \S \ref{sec:data}, following a comprehensive review of the literature and a detailed analysis of the posts in our dataset, we developed a taxonomy of lived drug experiences within online discussions. This taxonomy is structured across three levels: connection, subject, and objective. Through multiple iterations and evaluations, we refined our categorization to capture the nuances of online drug-related narratives accurately. Table \ref{tab:definition} presents the rationale, detailed definitions, and examples for each level, offering insights into the complexity and diversity of these experiences shared online.

\begin{table*}[h]
\centering
\resizebox{\linewidth}{!}{%
\begin{tabular}{@{}c|l|l|l@{}}
\toprule
\multicolumn{1}{c|}{\textbf{Dimension}} & \textbf{Rationale} & \textbf{Code} & \textbf{Definition and Example} \\ \hline
\multirow{2}{*}{Connection} & \multirow{2}{*}{\begin{tabular}[c]{@{}l@{}}What is the primary purpose of the post? \\ Is the post asking for something from \\the community or is it meant to share \\stories and lived experiences?\end{tabular}} & Disclosure & \begin{tabular}[c]{@{}l@{}}Making others aware of the activity related to drug use, both \\ primary and secondary accounts – including possession. \\ Example: ``I've been taking X for anxiety. I finally \\ went to a doctor who told me he wouldn't prescribe me \\ any benzo.''\end{tabular} \\ \cline{3-4} 
&  & Inquisition & \begin{tabular}[c]{@{}l@{}}Asking for questions and advice on SUD. \\ Example: ``What is your cure to prevent coke side effects \\ or things you do before, during, and after your session?''\end{tabular} \\ \hline
\multirow{4}{*}{Subject} & \multirow{4}{*}{What is the overarching subject of the post?} & Dependency & \begin{tabular}[c]{@{}l@{}}The medical term used to describe drug or alcohol use that continues \\ even when significant problems related to their use have developed.\\ Example: ``I'm 23 in usa been on drugs since I was a teen. drugs \\r bad but I chose to use, if you knew me know why!''\end{tabular} \\ \cline{3-4} 
&  & Recovery & \begin{tabular}[c]{@{}l@{}}Describing the process of overcoming substance use disorder and \\ regaining physical, emotional, and mental health. \\ Example: ``I am sober for six weeks and it is like my life \\ has changed completely.''\end{tabular} \\ \cline{3-4} 
&  & Other & \begin{tabular}[c]{@{}l@{}}Other types of drug discussion NOT related to a recovery experience \\ or indicative of Recovery nor Dependency. Can include general use. \\ Example: ``do you know of a drug that feels like X \\but wouldn't test +?'' \end{tabular} \\ \cline{3-4} 
&  & N/A & \begin{tabular}[c]{@{}l@{}} Unrelated discussions, NOT related to any form of substance use.\\ Example: ``have you listened to Zoo Band?'' \end{tabular}\\ \hline
\multirow{13}{*}{Objective} & \multirow{13}{*}{\begin{tabular}[c]{@{}l@{}}What are main themes or topic(s)\\ corresponding to subject of the post?\\ These themes provide information \\into users' motivation behind their post; \\ including reasons for taking substances, \\the desired effects users hope to achieve, \\the benefits of using the substance,\\asking for support and encouragement.\end{tabular}} & Combination of Substances & \begin{tabular}[c]{@{}l@{}}The use of two or more substances at the same time.\\ Example: ``I'm taking 10mg of meth and 10mg \\ X a day from the clinic. I'm on prescription,\\ 10mg dex a day but sometimes go over.''\end{tabular} \\ \cline{3-4} 
&  & Effects & \begin{tabular}[c]{@{}l@{}}Physical or emotional effect: The desired or undesired effects of the \\ substance on the user's body and mind. \\ Example: ``I use amps for anxiety but now I have a bad craving.''\end{tabular} \\ \cline{3-4} 
&  & Legality & \begin{tabular}[c]{@{}l@{}}The legal status of the substance in the user's jurisdiction. \\ Example: ``Is benzo legal in the US? Will I pass the test?''\end{tabular} \\ \cline{3-4} 
&  & Mental Health & \begin{tabular}[c]{@{}l@{}}The impact of the substance on the user's mental state, including \\ mood, cognition, and emotional regulation.\\ Example: ``I feel like I have no options. The only thing\\ kept me from feeling bad and having panic \\ attacks or episodes of depression for extended \\periods are xanas.''\end{tabular} \\ \cline{3-4} 
&  & Methods of Ingestion & \begin{tabular}[c]{@{}l@{}}The way in which the substance is consumed, such as smoking,\\ inhaling, injecting, or swallowing. This encompasses different\\ routes of administration and dosage-related considerations that \\play a crucial role in determining the substance's effects users. \\ Example: ``I'm smoking 3g coke, it's the way I like to use.''\end{tabular} \\ \cline{3-4} 
&  & Nurturant Support \& Morality & \begin{tabular}[c]{@{}l@{}}User's thoughts and feelings about the substance and their \\ own use of it, including feelings of guilt, shame, or self-judgment. \\ Nurturant support includes emotional, network, and esteem.\\ Example: ``I'm a horrible person because I am an addict, \\ and I can't quit.''\end{tabular} \\ \cline{3-4} 
&  & Overdose & \begin{tabular}[c]{@{}l@{}}The consumption of more of a substance than the body can \\ safely handle, resulting in serious health problems or death.\\ Example: ``I overdosed last week, it was the scariest \\ event of my life.''\end{tabular} \\ \cline{3-4} 
&  & Quality & \begin{tabular}[c]{@{}l@{}}The purity or potency of the substance, or the user's perception of it.\\ Example: ``Got some prams that I'msure are pure, \\same size/dimensions/and weight.'' \end{tabular} \\ \cline{3-4} 
&  & Relapse & \begin{tabular}[c]{@{}l@{}}The return to using a substance after a period of abstinence.\\ Example: ``I relapsed on opiates last week, but I'm ok and in \\ treatment and I want to stay sober.''\end{tabular} \\ \cline{3-4}  &  & Safety & \begin{tabular}[c]{@{}l@{}}The perceived or actual risk associated with using the substance, \\ includes risks of overdose, addiction, or other negative consequences.\\ Example: ``Can my ex purposefully switch \\ my syringe with one of hers because \\ she is pissed about a comment I made about not risking Hep C?''\end{tabular} \\ \cline{3-4} 
&  & Withdrawal & \begin{tabular}[c]{@{}l@{}}The physical and psychological symptoms that occur when a person \\ stops using a substance that they have been addicted to.\\ Example: ``I'm going through withdrawal now, it's awful.''\end{tabular} \\ \cline{3-4} 
&  & Other & \begin{tabular}[c]{@{}l@{}}Any other objective that is not covered by the above categories.\\ Example: ``There was a post on various components of\\ xanax and the meaning, it broke names into their \\various components.''
\end{tabular} \\ \cline{3-4} 
&  & N/A & \begin{tabular}[c]{@{}l@{}}Not related to SUD objective.\\ Example: ``Want to chat, hit me up.'' \end{tabular}\\ \hline
\end{tabular}%
}
\caption{Taxonomy of Lived
Drug Experiences in Online Discussions}
\label{tab:definition}
\end{table*}

\section{Details of Classification Procedure and Results} \label{app:full_result}
Table \ref{tab:full-results} presents the comprehensive list of classifiers and the corresponding results we obtained for the three levels of classification performed. The table encompasses a detailed comparison across different metrics, providing insights into the performance of each classifier within the context of our study. 
The classifications were conducted across three levels. 
For the baseline models, for both Connection and Subject levels, LogR and SVM models employ a one-versus-all strategy, whereas the KNN and RF models directly support multi-class classification. At the Objective level, however, we adopt the `MultiOutputClassifier' strategy to address the multi-label classification task.

For the Objective level in transformer-based models, we adopted a multi-label classification strategy to manage samples that simultaneously belong to multiple categories.

The number of parameters in The BERT model is 110 million, while the RoBERTa and DeBERTa models have 125 million and 276 million parameters, respectively. We used the T4 GPU provided by Google Colab for our transformer-based models. 
For SentBERT model, we used seven few-shot samples from each category at both the Connection and Subject levels, and only four examples per category at the Objective level.

\begin{table*}[h!]
\centering
\resizebox{\linewidth}{!}{%
\begin{tabular}{@{}lllllllllllll@{}}
\toprule
\multicolumn{3}{c}{\texttt{\textbf{}}} & \multicolumn{3}{c}{\texttt{\textbf{Connection}}} & \multicolumn{3}{c}{\texttt{\textbf{Subject}}} & \multicolumn{3}{c}{\texttt{\textbf{Objective}}}\\ 
 \cmidrule(lr){4-6} \cmidrule(lr){7-9} \cmidrule(lr){10-12}
\texttt{\textbf{}} & \texttt{\textbf{Feature}} & \texttt{\textbf{Classifiers}} & \texttt{\textbf{Prec}} & \texttt{\textbf{Rec}} & \texttt{\textbf{F1}} & \texttt{\textbf{Prec}} & \texttt{\textbf{Rec}} & \texttt{\textbf{F1}} & \texttt{\textbf{Prec}} & \texttt{\textbf{Rec}} & \texttt{\textbf{F1}}\\ 
\cmidrule(lr){1-3}\cmidrule(lr){4-6} \cmidrule(l){7-9} \cmidrule(l){10-12}

\multirow{12}{*}{\texttt{Baseline}} & \texttt{TF-IDF} & \texttt{KNN} & 0.77 & 0.72 & 0.68 & 0.49 & 0.51 & 0.49 & 0.35 & 0.44 & 0.39\\
& \texttt{TF-IDF} & \texttt{SVM} & 0.76 & 0.60 & 0.47 & 0.65 & 0.70 & 0.66 & 0.35 & 0.47 & 0.41\\
& \texttt{TF-IDF} & \texttt{RF} & 0.67 & 0.66 & 0.62 & 0.52 & 0.58 & 0.55 & 0.37 & 0.32 & 0.35\\
& \texttt{TF-IDF} & \texttt{LogR} & 0.76 & 0.74 & 0.72 & 0.58 & 0.53 & 0.52 & 0.46 & 0.47 & 0.41\\
 & \texttt{BERT} & \texttt{LogR} & 0.71 & 0.72 & 0.71 & 0.53 & 0.58 & 0.55 & 0.60 & 0.47 & 0.49\\
 & \texttt{BERT} & \texttt{KNN} & 0.67 & 0.62 & 0.53 & 0.55 & 0.53 & 0.52 & 0.36 & 0.45 & 0.40\\
 & \texttt{BERT} & \texttt{RF} & 0.39 & 0.51 & 0.42 & 0.55 & 0.60 & 0.58 & 0.45 & 0.39 & 0.40\\
 & \texttt{BERT} & \texttt{SVM} & 0.75 & 0.75 & 0.75 & 0.56 & 0.60 & 0.58 & 0.52 & 0.45 & 0.44\\ 
& \texttt{ada-002} & \texttt{SVM} & 0.8 & 0.8 & 0.8 & 0.26 & 0.53 & 0.35 & 0.33 & 0.29 & 0.30\\
 & \texttt{ada-002} & \texttt{LogR} & 0.82 & 0.82 & 0.81 & 0.57 & 0.63 & 0.58 & 0.48 & 0.30 & 0.34\\
& \texttt{ada-002} & \texttt{RF} & 0.64 & 0.62 & 0.62 & 0.46 & 0.57 & 0.51 & 0.39 & 0.23 & 0.26\\
& \texttt{ada-002} & \texttt{KNN} & 0.78 & 0.78 & 0.78 & 0.46 & 0.57 & 0.51 & 0.39 & 0.23 & 0.26\\
 \cmidrule(lr){1-3}
\multirow{4}{*}{\texttt{Transformer-Based}} & \texttt{BERT} & \texttt{BERT} & 0.78 & 0.68 & 0.73 & 0.58 & 0.56 & 0.54 & 0.38 & 0.41 & 0.41\\
& \texttt{DeBERTa} & \texttt{DeBERTa} & \textbf{0.95} & 0.87 & 0.90 & 0.66 & 0.71 & 0.69 & 0.51 & 0.44 & 0.43\\
& \texttt{RoBERTa} & \texttt{RoBERTa} & 0.85 & 0.81 & 0.83 & 0.63 & 0.69 & 0.66 & 0.39 & 0.43 & 0.41\\
 & \texttt{BioBERT} & \texttt{BioBERT} & 0.54 & 0.9 & 0.67 & 0.59 & 0.62 & 0.60 & 0.38 & 0.45 & 0.41\\ 
\texttt{} & - & \texttt{SetFit} & 0.90 & 0.64 & 0.75 & 0.62 & 0.39 & 0.45 & 0.37 & 0.43 & 0.39\\ 
\cmidrule(lr){1-3}
\multirow{9}{*}{\texttt{LLM}} & - & \texttt{GPT3.5 I} & 0.6 & 0.6 & 0.6 & 0.73 & 0.45 & 0.4 & 0.57 & 0.61 & 0.56 \\
& - & \texttt{GPT3.5 I + D} & 0.49 & 0.43 & 0.39 & 0.66 & 0.47 & 0.42 & 0.54 & 0.57 & 0.53 \\
& - & \texttt{GPT3.5 I + D + E} & 0.71 & 0.7 & 0.69 & 0.75 & 0.68 & 0.68 & 0.61 & 0.69 & 0.62 \\
& - & \texttt{GPT4 I} & 0.73 & 0.57 & 0.53 & 0.74 & 0.34 & 0.23 & 0.54 & 0.74 & 0.6 \\
& - & \texttt{GPT4 I + D} & 0.76 & 0.45 & 0.32 & \textbf{0.77} & 0.64 & 0.66 & 0.56 & \textbf{0.76} & 0.61 \\
 & - & \texttt{GPT4 I + D + E} & 0.91 & \textbf{0.91} & \textbf{0.91} & 0.74 & \textbf{0.72} & \textbf{0.73} & \textbf{0.71} & 0.66 & \textbf{0.64}\\
  & - & \texttt{Mixtral I} & 0.34 & 0.58 & 0.43 & 0.1 & 0.32 & 0.16 & 0.4 & 0.68 & 0.48\\
& - & \texttt{Mixtral I + D} & 0.17 & 0.42 & 0.24 & 0.59 & 0.43 & 0.37 & 0.54 & 0.68 & 0.55\\
& - & \texttt{Mixtral I + D + E} & 0.78 & 0.51 & 0.43 & 0.72 & 0.34 & 0.23 & 0.42 & 0.84 & 0.53\\
\bottomrule
\end{tabular}%
}
\caption{Weighted Precision, Recall, and F1 for all classification models.}
\label{tab:full-results}
\end{table*}

\section{Multi-level, Multi-class Frequencies}
Table \ref{tab:cross-tab} presents the distribution frequencies for each objective, categorized by connection and subject types. These results illustrate how the various classes are represented across a dataset of 1,500 entries, which includes 1,000 posts annotated using our best-performing model, complemented by an additional 500 manually annotated data points. This comprehensive overview aids in understanding the prevalence and patterns of different objectives within the context of the study, highlighting the robustness and coverage of our annotation approach.

\begin{table*}[h]
\centering
\resizebox{0.8\textwidth}{!}{%
\begin{tabular}{@{}cccccccccccccc@{}}
\toprule
\rotatebox{90}{\texttt{}} &
\rotatebox{90}{\texttt{}} &
\rotatebox{90}{\texttt{Comb. of Substances}} &
\rotatebox{90}{\texttt{Effects}} &
\rotatebox{90}{\texttt{Legality}} &
\rotatebox{90}{\texttt{Mental Health}} &
\rotatebox{90}{\texttt{Methods of Ingestion}} &
\rotatebox{90}{\texttt{Nurturant \& Morality}} &
\rotatebox{90}{\texttt{Overdose}} &
\rotatebox{90}{\texttt{Quality}} &
\rotatebox{90}{\texttt{Relapse}} &
\rotatebox{90}{\texttt{Safety}} &
\rotatebox{90}{\texttt{Withdrawal}} &
\rotatebox{90}{\texttt{Total}} \\ 
\cmidrule(l){3-14}
\multirow{3}{*}{\texttt{Inquisition}} & \texttt{Dependency} & 27 & \textbf{88} & 3 & 34 & 61 & 19 & 11 & 12 & 14 & 32 & 26 & 328 \\
                                      & \texttt{Recovery}   & 2  & 9           & 1 & 7  & 5  & \textbf{17} & 1 & 1 & 15 & 3  & \textbf{17} & 78 \\
                                      & \texttt{Other}      & 31 & \textbf{119}& 18& 28 & 87 & 18          & 8 & 20& 4  & 40 & 3           & 397 \\ \cmidrule(l){1-2}
\multirow{3}{*}{\texttt{Disclosure}}  & \texttt{Dependency} & 34 & 84          & 9 & 28 & \textbf{89} & 11 & 5 & 9 & 13 & 38 & 50          & 370 \\
                                      & \texttt{Recovery}   & 5  & 8           & 1 & 11 & 10          & 4  & 4 & 0 & 13 & 12 & \textbf{23} & 91 \\ 
                                      & \texttt{Other}      & 78 & 200         & 23& 29 & \textbf{219}& 9  & 6 & 43& 1  & 109& 10          & 732 \\\cmidrule(l){1-2}
\multicolumn{2}{c}{\texttt{Total}}    & 177& 508            & 55& 137& 471     & 78 & 35& 85& 60 & 234& 129         & \\
\bottomrule
\end{tabular}%
}
\caption{Frequencies of each objective with respect to each connection and lived experience types. The 'N/A' label for lived experience and 'N/A' and "Other' objectives labels are excluded from this table.}
\label{tab:cross-tab}
\end{table*}

\section{LIWC and Statistical Analysis} \label{app:LIWC}

Table \ref{tab:LIWC} shows the ratios of mean scores for posts labeled as Recovery and Dependency ($\mu_{Recovery}/\mu_{Dependency}$) %across various LIWC categories, while Table \ref{tab:liwc2} presents 
and the ratios of mean scores for Inquisition and Disclosure ($\mu_{Inquisition}/\mu_{Dependency}$) across different LIWC categories. LIWC assesses the text by calculating the proportion of words belonging to different psychologically relevant categories.

\begin{table*}[htbp]
% \centering
% \resizebox{0.6\textwidth}{!}{%
\centering
\resizebox{0.6\textwidth}{!}{%
\begin{tabular}{@{}lccccccccccc|c@{}}
\toprule
 & \multicolumn{11}{c|}{\texttt{$\mu_{Inquisition}/\mu_{Disclosure}$}} & \texttt{$\mu_{Recovery}/\mu_{Dependency}$} \\ \cmidrule(lr){2-12} \cmidrule(lr){13-13}
\rotatebox{90}{\texttt{LWC}} & \rotatebox{90}{\texttt{Recovery}} & \rotatebox{90}{\texttt{Dependency}} & \rotatebox{90}{\texttt{Effects}} & \rotatebox{90}{\texttt{Methods of Ingestion}} & \rotatebox{90}{\texttt{Comb. of Substances}} & \rotatebox{90}{\texttt{Mental Health}} & \rotatebox{90}{\texttt{Nurturant \& Morality}} & \rotatebox{90}{\texttt{Withdrawal}} & \rotatebox{90}{\texttt{Safety}} & \rotatebox{90}{\texttt{Relapse}} & \rotatebox{90}{\texttt{ALL}} &\rotatebox{90}{\texttt{Effects}} \\ \midrule
\textbf{achieve} & \cellcolor[HTML]{FFFFFF} & \cellcolor[HTML]{FFFFFF} & \cellcolor[HTML]{FFFFFF} & \cellcolor[HTML]{F5DCDC}1.18 & \cellcolor[HTML]{DDEBF7}0.95 & \cellcolor[HTML]{FFFFFF} & \cellcolor[HTML]{FFFFFF} & \cellcolor[HTML]{FFFFFF} & \cellcolor[HTML]{FFFFFF} & \cellcolor[HTML]{FFFFFF} & \cellcolor[HTML]{FFFFFF} & \cellcolor[HTML]{FFFFFF} \\
\textbf{acquire} & \cellcolor[HTML]{FFFFFF} & \cellcolor[HTML]{E6B5B6}1.38 & \cellcolor[HTML]{D65C5E}1.68 & \cellcolor[HTML]{D65C5E}1.92 & \cellcolor[HTML]{D65C5E}3.08 & \cellcolor[HTML]{D65C5E}1.75 & \cellcolor[HTML]{FFFFFF} & \cellcolor[HTML]{FFFFFF} & \cellcolor[HTML]{D65C5E}3.16 & \cellcolor[HTML]{FFFFFF} & \cellcolor[HTML]{E6B5B6}1.35 & \cellcolor[HTML]{FFFFFF} \\
\textbf{affect} & \cellcolor[HTML]{FFFFFF} & \cellcolor[HTML]{5B9BD5}0.54 & \cellcolor[HTML]{FFFFFF} & \cellcolor[HTML]{FFFFFF} & \cellcolor[HTML]{FFFFFF} & \cellcolor[HTML]{FFFFFF} & \cellcolor[HTML]{FFFFFF} & \cellcolor[HTML]{5B9BD5}0.69 & \cellcolor[HTML]{FFFFFF} & \cellcolor[HTML]{FFFFFF} & \cellcolor[HTML]{5B9BD5}0.67 & \cellcolor[HTML]{FFFFFF} \\
\textbf{affiliation} & \cellcolor[HTML]{FFFFFF} & \cellcolor[HTML]{FFFFFF} & \cellcolor[HTML]{FFFFFF} & \cellcolor[HTML]{FFFFFF} & \cellcolor[HTML]{FFFFFF} & \cellcolor[HTML]{FFFFFF} & \cellcolor[HTML]{FFFFFF} & \cellcolor[HTML]{5B9BD5}0.45 & \cellcolor[HTML]{FFFFFF} & \cellcolor[HTML]{FFFFFF} & \cellcolor[HTML]{FFFFFF} & \cellcolor[HTML]{FFFFFF} \\
\textbf{allnone} & \cellcolor[HTML]{FFFFFF} & \cellcolor[HTML]{FFFFFF} & \cellcolor[HTML]{FFFFFF} & \cellcolor[HTML]{FFFFFF} & \cellcolor[HTML]{FFFFFF} & \cellcolor[HTML]{FFFFFF} & \cellcolor[HTML]{FFFFFF} & \cellcolor[HTML]{5B9BD5}0.59 & \cellcolor[HTML]{FFFFFF} & \cellcolor[HTML]{FFFFFF} & \cellcolor[HTML]{DDEBF7}0.75 & \cellcolor[HTML]{FFFFFF} \\
\textbf{allure} & \cellcolor[HTML]{FFFFFF} & \cellcolor[HTML]{FFFFFF} & \cellcolor[HTML]{FFFFFF} & \cellcolor[HTML]{DDEBF7}0.90 & \cellcolor[HTML]{DDEBF7}0.89 & \cellcolor[HTML]{FFFFFF} & \cellcolor[HTML]{FFFFFF} & \cellcolor[HTML]{FFFFFF} & \cellcolor[HTML]{FFFFFF} & \cellcolor[HTML]{FFFFFF} & \cellcolor[HTML]{FFFFFF} & \cellcolor[HTML]{FFFFFF} \\
\textbf{analytic} & \cellcolor[HTML]{FFFFFF} & \cellcolor[HTML]{FFFFFF} & \cellcolor[HTML]{DDEBF7}0.78 & \cellcolor[HTML]{DDEBF7}0.76 & \cellcolor[HTML]{DDEBF7}0.77 & \cellcolor[HTML]{FFFFFF} & \cellcolor[HTML]{FFFFFF} & \cellcolor[HTML]{FFFFFF} & \cellcolor[HTML]{DDEBF7}0.75 & \cellcolor[HTML]{FFFFFF} & \cellcolor[HTML]{DDEBF7}0.85 & \cellcolor[HTML]{FFFFFF} \\
\textbf{authentic} & \cellcolor[HTML]{FFFFFF} & \cellcolor[HTML]{FFFFFF} & \cellcolor[HTML]{FFFFFF} & \cellcolor[HTML]{F5DCDC}1.17 & \cellcolor[HTML]{FFFFFF} & \cellcolor[HTML]{FFFFFF} & \cellcolor[HTML]{FFFFFF} & \cellcolor[HTML]{FFFFFF} & \cellcolor[HTML]{FFFFFF} & \cellcolor[HTML]{FFFFFF} & \cellcolor[HTML]{FFFFFF} & \cellcolor[HTML]{FFFFFF} \\
\textbf{cause} & \cellcolor[HTML]{FFFFFF} & \cellcolor[HTML]{D65C5E}1.82 & \cellcolor[HTML]{E6B5B6}1.40 & \cellcolor[HTML]{D65C5E}1.54 & \cellcolor[HTML]{D65C5E}1.73 & \cellcolor[HTML]{F5DCDC}1.20 & \cellcolor[HTML]{FFFFFF} & \cellcolor[HTML]{FFFFFF} & \cellcolor[HTML]{E6B5B6}1.47 & \cellcolor[HTML]{FFFFFF} & \cellcolor[HTML]{E6B5B6}1.51 & \cellcolor[HTML]{FFFFFF} \\
\textbf{certitude} & \cellcolor[HTML]{FFFFFF} & \cellcolor[HTML]{FFFFFF} & \cellcolor[HTML]{FFFFFF} & \cellcolor[HTML]{FFFFFF} & \cellcolor[HTML]{FFFFFF} & \cellcolor[HTML]{FFFFFF} & \cellcolor[HTML]{FFFFFF} & \cellcolor[HTML]{FFFFFF} & \cellcolor[HTML]{FFFFFF} & \cellcolor[HTML]{FFFFFF} & \cellcolor[HTML]{5B9BD5}0.69 & \cellcolor[HTML]{FFFFFF} \\
\textbf{clout} & \cellcolor[HTML]{FFFFFF} & \cellcolor[HTML]{FFFFFF} & \cellcolor[HTML]{DDEBF7}0.74 & \cellcolor[HTML]{5B9BD5}0.69 & \cellcolor[HTML]{DDEBF7}0.71 & \cellcolor[HTML]{FFFFFF} & \cellcolor[HTML]{FFFFFF} & \cellcolor[HTML]{FFFFFF} & \cellcolor[HTML]{5B9BD5}0.54 & \cellcolor[HTML]{FFFFFF} & \cellcolor[HTML]{DDEBF7}0.90 & \cellcolor[HTML]{FFFFFF} \\
\textbf{cognition} & \cellcolor[HTML]{FFFFFF} & \cellcolor[HTML]{F5DCDC}1.27 & \cellcolor[HTML]{F5DCDC}1.26 & \cellcolor[HTML]{E6B5B6}1.36 & \cellcolor[HTML]{E6B5B6}1.31 & \cellcolor[HTML]{FFFFFF} & \cellcolor[HTML]{E6B5B6}1.31 & \cellcolor[HTML]{FFFFFF} & \cellcolor[HTML]{F5DCDC}1.25 & \cellcolor[HTML]{FFFFFF} & \cellcolor[HTML]{E6B5B6}1.33 & \cellcolor[HTML]{FFFFFF} \\
\textbf{cogproc} & \cellcolor[HTML]{FFFFFF} & \cellcolor[HTML]{E6B5B6}1.32 & \cellcolor[HTML]{E6B5B6}1.29 & \cellcolor[HTML]{E6B5B6}1.41 & \cellcolor[HTML]{E6B5B6}1.35 & \cellcolor[HTML]{F5DCDC}1.24 & \cellcolor[HTML]{E6B5B6}1.37 & \cellcolor[HTML]{FFFFFF} & \cellcolor[HTML]{E6B5B6}1.28 & \cellcolor[HTML]{FFFFFF} & \cellcolor[HTML]{E6B5B6}1.42 & \cellcolor[HTML]{FFFFFF} \\
\textbf{comm} & \cellcolor[HTML]{FFFFFF} & \cellcolor[HTML]{D65C5E}1.53 & \cellcolor[HTML]{DDEBF7}0.98 & \cellcolor[HTML]{F5DCDC}1.25 & \cellcolor[HTML]{DDEBF7}0.78 & \cellcolor[HTML]{FFFFFF} & \cellcolor[HTML]{FFFFFF} & \cellcolor[HTML]{FFFFFF} & \cellcolor[HTML]{DDEBF7}0.96 & \cellcolor[HTML]{FFFFFF} & \cellcolor[HTML]{E6B5B6}1.27 & \cellcolor[HTML]{FFFFFF} \\
\textbf{conflict} & \cellcolor[HTML]{FFFFFF} & \cellcolor[HTML]{FFFFFF} & \cellcolor[HTML]{F5DCDC}1.21 & \cellcolor[HTML]{FFFFFF} & \cellcolor[HTML]{FFFFFF} & \cellcolor[HTML]{FFFFFF} & \cellcolor[HTML]{FFFFFF} & \cellcolor[HTML]{5B9BD5}0.39 & \cellcolor[HTML]{FFFFFF} & \cellcolor[HTML]{FFFFFF} & \cellcolor[HTML]{5B9BD5}0.66 & \cellcolor[HTML]{FFFFFF} \\
\textbf{curiosity} & \cellcolor[HTML]{FFFFFF} & \cellcolor[HTML]{FFFFFF} & \cellcolor[HTML]{FFFFFF} & \cellcolor[HTML]{E6B5B6}1.33 & \cellcolor[HTML]{FFFFFF} & \cellcolor[HTML]{FFFFFF} & \cellcolor[HTML]{FFFFFF} & \cellcolor[HTML]{FFFFFF} & \cellcolor[HTML]{E6B5B6}1.39 & \cellcolor[HTML]{FFFFFF} & \cellcolor[HTML]{D65C5E}2.02 & \cellcolor[HTML]{FFFFFF} \\
\textbf{differ} & \cellcolor[HTML]{FFFFFF} & \cellcolor[HTML]{FFFFFF} & \cellcolor[HTML]{E6B5B6}1.33 & \cellcolor[HTML]{E6B5B6}1.48 & \cellcolor[HTML]{E6B5B6}1.42 & \cellcolor[HTML]{FFFFFF} & \cellcolor[HTML]{FFFFFF} & \cellcolor[HTML]{FFFFFF} & \cellcolor[HTML]{E6B5B6}1.31 & \cellcolor[HTML]{FFFFFF} & \cellcolor[HTML]{E6B5B6}1.33 & \cellcolor[HTML]{FFFFFF} \\
\textbf{discrep} & \cellcolor[HTML]{FFFFFF} & \cellcolor[HTML]{E6B5B6}1.39 & \cellcolor[HTML]{E6B5B6}1.27 & \cellcolor[HTML]{D65C5E}1.62 & \cellcolor[HTML]{D65C5E}1.87 & \cellcolor[HTML]{D65C5E}1.74 & \cellcolor[HTML]{FFFFFF} & \cellcolor[HTML]{FFFFFF} & \cellcolor[HTML]{E6B5B6}1.52 & \cellcolor[HTML]{FFFFFF} & \cellcolor[HTML]{D65C5E}1.59 & \cellcolor[HTML]{FFFFFF} \\
\textbf{drives} & \cellcolor[HTML]{FFFFFF} & \cellcolor[HTML]{FFFFFF} & \cellcolor[HTML]{FFFFFF} & \cellcolor[HTML]{DDEBF7}0.87 & \cellcolor[HTML]{DDEBF7}0.75 & \cellcolor[HTML]{FFFFFF} & \cellcolor[HTML]{FFFFFF} & \cellcolor[HTML]{FFFFFF} & \cellcolor[HTML]{FFFFFF} & \cellcolor[HTML]{FFFFFF} & \cellcolor[HTML]{FFFFFF} & \cellcolor[HTML]{FFFFFF} \\
\textbf{emo\_anger} & \cellcolor[HTML]{FFFFFF} & \cellcolor[HTML]{FFFFFF} & \cellcolor[HTML]{5B9BD5}0.60 & \cellcolor[HTML]{FFFFFF} & \cellcolor[HTML]{FFFFFF} & \cellcolor[HTML]{FFFFFF} & \cellcolor[HTML]{FFFFFF} & \cellcolor[HTML]{FFFFFF} & \cellcolor[HTML]{FFFFFF} & \cellcolor[HTML]{FFFFFF} & \cellcolor[HTML]{DDEBF7}0.70 & \cellcolor[HTML]{FFFFFF} \\
\textbf{emo\_neg} & \cellcolor[HTML]{FFFFFF} & \cellcolor[HTML]{5B9BD5}0.66 & \cellcolor[HTML]{FFFFFF} & \cellcolor[HTML]{FFFFFF} & \cellcolor[HTML]{E6B5B6}1.35 & \cellcolor[HTML]{FFFFFF} & \cellcolor[HTML]{FFFFFF} & \cellcolor[HTML]{FFFFFF} & \cellcolor[HTML]{E6B5B6}1.37 & \cellcolor[HTML]{FFFFFF} & \cellcolor[HTML]{FFFFFF} & \cellcolor[HTML]{FFFFFF} \\
\textbf{emo\_pos} & \cellcolor[HTML]{FFFFFF} & \cellcolor[HTML]{FFFFFF} & \cellcolor[HTML]{5B9BD5}0.38 & \cellcolor[HTML]{FFFFFF} & \cellcolor[HTML]{FFFFFF} & \cellcolor[HTML]{FFFFFF} & \cellcolor[HTML]{FFFFFF} & \cellcolor[HTML]{FFFFFF} & \cellcolor[HTML]{FFFFFF} & \cellcolor[HTML]{FFFFFF} & \cellcolor[HTML]{5B9BD5}0.41 & \cellcolor[HTML]{FFFFFF} \\
\textbf{emo\_sad} & \cellcolor[HTML]{FFFFFF} & \cellcolor[HTML]{FFFFFF} & \cellcolor[HTML]{FFFFFF} & \cellcolor[HTML]{FFFFFF} & \cellcolor[HTML]{FFFFFF} & \cellcolor[HTML]{FFFFFF} & \cellcolor[HTML]{FFFFFF} & \cellcolor[HTML]{FFFFFF} & \cellcolor[HTML]{FFFFFF} & \cellcolor[HTML]{FFFFFF} & \cellcolor[HTML]{DDEBF7}0.86 & \cellcolor[HTML]{D65C5E}3.43 \\
\textbf{emotion} & \cellcolor[HTML]{FFFFFF} & \cellcolor[HTML]{5B9BD5}0.60 & \cellcolor[HTML]{FFFFFF} & \cellcolor[HTML]{FFFFFF} & \cellcolor[HTML]{FFFFFF} & \cellcolor[HTML]{FFFFFF} & \cellcolor[HTML]{FFFFFF} & \cellcolor[HTML]{5B9BD5}0.60 & \cellcolor[HTML]{FFFFFF} & \cellcolor[HTML]{FFFFFF} & \cellcolor[HTML]{5B9BD5}0.60 & \cellcolor[HTML]{FFFFFF} \\
\textbf{family} & \cellcolor[HTML]{FFFFFF} & \cellcolor[HTML]{DDEBF7}0.72 & \cellcolor[HTML]{5B9BD5}0.60 & \cellcolor[HTML]{5B9BD5}0.60 & \cellcolor[HTML]{FFFFFF} & \cellcolor[HTML]{FFFFFF} & \cellcolor[HTML]{FFFFFF} & \cellcolor[HTML]{5B9BD5}0.57 & \cellcolor[HTML]{FFFFFF} & \cellcolor[HTML]{FFFFFF} & \cellcolor[HTML]{5B9BD5}0.47 & \cellcolor[HTML]{E6B5B6}1.49 \\
\textbf{fatigue} & \cellcolor[HTML]{FFFFFF} & \cellcolor[HTML]{FFFFFF} & \cellcolor[HTML]{FFFFFF} & \cellcolor[HTML]{FFFFFF} & \cellcolor[HTML]{FFFFFF} & \cellcolor[HTML]{FFFFFF} & \cellcolor[HTML]{FFFFFF} & \cellcolor[HTML]{FFFFFF} & \cellcolor[HTML]{FFFFFF} & \cellcolor[HTML]{FFFFFF} & \cellcolor[HTML]{DDEBF7}0.82 & \cellcolor[HTML]{FFFFFF} \\
\textbf{feeling} & \cellcolor[HTML]{FFFFFF} & \cellcolor[HTML]{FFFFFF} & \cellcolor[HTML]{FFFFFF} & \cellcolor[HTML]{FFFFFF} & \cellcolor[HTML]{D65C5E}2.56 & \cellcolor[HTML]{FFFFFF} & \cellcolor[HTML]{FFFFFF} & \cellcolor[HTML]{FFFFFF} & \cellcolor[HTML]{FFFFFF} & \cellcolor[HTML]{FFFFFF} & \cellcolor[HTML]{FFFFFF} & \cellcolor[HTML]{FFFFFF} \\
\textbf{female} & \cellcolor[HTML]{FFFFFF} & \cellcolor[HTML]{FFFFFF} & \cellcolor[HTML]{5B9BD5}0.67 & \cellcolor[HTML]{FFFFFF} & \cellcolor[HTML]{FFFFFF} & \cellcolor[HTML]{FFFFFF} & \cellcolor[HTML]{FFFFFF} & \cellcolor[HTML]{5B9BD5}0.17 & \cellcolor[HTML]{FFFFFF} & \cellcolor[HTML]{FFFFFF} & \cellcolor[HTML]{DDEBF7}0.77 & \cellcolor[HTML]{FFFFFF} \\
\textbf{focusfuture} & \cellcolor[HTML]{FFFFFF} & \cellcolor[HTML]{FFFFFF} & \cellcolor[HTML]{F5DCDC}1.20 & \cellcolor[HTML]{DDEBF7}0.99 & \cellcolor[HTML]{E6B5B6}1.43 & \cellcolor[HTML]{FFFFFF} & \cellcolor[HTML]{FFFFFF} & \cellcolor[HTML]{FFFFFF} & \cellcolor[HTML]{D65C5E}1.57 & \cellcolor[HTML]{FFFFFF} & \cellcolor[HTML]{FFFFFF} & \cellcolor[HTML]{FFFFFF} \\
\textbf{focuspast} & \cellcolor[HTML]{FFFFFF} & \cellcolor[HTML]{FFFFFF} & \cellcolor[HTML]{F5DCDC}1.05 & \cellcolor[HTML]{F5DCDC}1.13 & \cellcolor[HTML]{F5DCDC}1.14 & \cellcolor[HTML]{FFFFFF} & \cellcolor[HTML]{FFFFFF} & \cellcolor[HTML]{FFFFFF} & \cellcolor[HTML]{F5DCDC}1.10 & \cellcolor[HTML]{FFFFFF} & \cellcolor[HTML]{FFFFFF} & \cellcolor[HTML]{FFFFFF} \\
\textbf{focuspresent} & \cellcolor[HTML]{FFFFFF} & \cellcolor[HTML]{FFFFFF} & \cellcolor[HTML]{F5DCDC}1.08 & \cellcolor[HTML]{F5DCDC}1.13 & \cellcolor[HTML]{F5DCDC}1.06 & \cellcolor[HTML]{FFFFFF} & \cellcolor[HTML]{E6B5B6}1.51 & \cellcolor[HTML]{FFFFFF} & \cellcolor[HTML]{F5DCDC}1.21 & \cellcolor[HTML]{FFFFFF} & \cellcolor[HTML]{F5DCDC}1.13 & \cellcolor[HTML]{FFFFFF} \\
\textbf{food} & \cellcolor[HTML]{FFFFFF} & \cellcolor[HTML]{FFFFFF} & \cellcolor[HTML]{FFFFFF} & \cellcolor[HTML]{FFFFFF} & \cellcolor[HTML]{DDEBF7}0.78 & \cellcolor[HTML]{FFFFFF} & \cellcolor[HTML]{FFFFFF} & \cellcolor[HTML]{FFFFFF} & \cellcolor[HTML]{FFFFFF} & \cellcolor[HTML]{FFFFFF} & \cellcolor[HTML]{FFFFFF} & \cellcolor[HTML]{FFFFFF} \\
\textbf{friend} & \cellcolor[HTML]{FFFFFF} & \cellcolor[HTML]{FFFFFF} & \cellcolor[HTML]{DDEBF7}0.98 & \cellcolor[HTML]{FFFFFF} & \cellcolor[HTML]{FFFFFF} & \cellcolor[HTML]{FFFFFF} & \cellcolor[HTML]{FFFFFF} & \cellcolor[HTML]{5B9BD5}0.24 & \cellcolor[HTML]{FFFFFF} & \cellcolor[HTML]{FFFFFF} & \cellcolor[HTML]{DDEBF7}0.76 & \cellcolor[HTML]{FFFFFF} \\
\textbf{health} & \cellcolor[HTML]{FFFFFF} & \cellcolor[HTML]{FFFFFF} & \cellcolor[HTML]{E6B5B6}1.47 & \cellcolor[HTML]{F5DCDC}1.25 & \cellcolor[HTML]{D65C5E}2.41 & \cellcolor[HTML]{E6B5B6}1.43 & \cellcolor[HTML]{FFFFFF} & \cellcolor[HTML]{FFFFFF} & \cellcolor[HTML]{D65C5E}2.38 & \cellcolor[HTML]{FFFFFF} & \cellcolor[HTML]{F5DCDC}1.26 & \cellcolor[HTML]{FFFFFF} \\
\textbf{home} & \cellcolor[HTML]{FFFFFF} & \cellcolor[HTML]{FFFFFF} & \cellcolor[HTML]{5B9BD5}0.23 & \cellcolor[HTML]{5B9BD5}0.25 & \cellcolor[HTML]{5B9BD5}0.09 & \cellcolor[HTML]{FFFFFF} & \cellcolor[HTML]{FFFFFF} & \cellcolor[HTML]{5B9BD5}0.12 & \cellcolor[HTML]{FFFFFF} & \cellcolor[HTML]{FFFFFF} & \cellcolor[HTML]{5B9BD5}0.50 & \cellcolor[HTML]{D65C5E}2.92 \\
\textbf{I} & \cellcolor[HTML]{FFFFFF} & \cellcolor[HTML]{FFFFFF} & \cellcolor[HTML]{F5DCDC}1.11 & \cellcolor[HTML]{F5DCDC}1.22 & \cellcolor[HTML]{E6B5B6}1.31 & \cellcolor[HTML]{FFFFFF} & \cellcolor[HTML]{FFFFFF} & \cellcolor[HTML]{FFFFFF} & \cellcolor[HTML]{E6B5B6}1.38 & \cellcolor[HTML]{FFFFFF} & \cellcolor[HTML]{FFFFFF} & \cellcolor[HTML]{FFFFFF} \\
\textbf{illness} & \cellcolor[HTML]{FFFFFF} & \cellcolor[HTML]{FFFFFF} & \cellcolor[HTML]{D65C5E}3.03 & \cellcolor[HTML]{FFFFFF} & \cellcolor[HTML]{D65C5E}3.03 & \cellcolor[HTML]{FFFFFF} & \cellcolor[HTML]{FFFFFF} & \cellcolor[HTML]{FFFFFF} & \cellcolor[HTML]{D65C5E}2.97 & \cellcolor[HTML]{FFFFFF} & \cellcolor[HTML]{FFFFFF} & \cellcolor[HTML]{FFFFFF} \\
\textbf{insight} & \cellcolor[HTML]{FFFFFF} & \cellcolor[HTML]{E6B5B6}1.42 & \cellcolor[HTML]{F5DCDC}1.12 & \cellcolor[HTML]{F5DCDC}1.21 & \cellcolor[HTML]{DDEBF7}0.93 & \cellcolor[HTML]{F5DCDC}1.25 & \cellcolor[HTML]{FFFFFF} & \cellcolor[HTML]{FFFFFF} & \cellcolor[HTML]{DDEBF7}0.99 & \cellcolor[HTML]{FFFFFF} & \cellcolor[HTML]{E6B5B6}1.48 & \cellcolor[HTML]{FFFFFF} \\
\textbf{ipron} & \cellcolor[HTML]{FFFFFF} & \cellcolor[HTML]{FFFFFF} & \cellcolor[HTML]{F5DCDC}1.06 & \cellcolor[HTML]{F5DCDC}1.07 & \cellcolor[HTML]{DDEBF7}0.96 & \cellcolor[HTML]{FFFFFF} & \cellcolor[HTML]{FFFFFF} & \cellcolor[HTML]{FFFFFF} & \cellcolor[HTML]{DDEBF7}0.99 & \cellcolor[HTML]{FFFFFF} & \cellcolor[HTML]{F5DCDC}1.20 & \cellcolor[HTML]{FFFFFF} \\
\textbf{leisure} & \cellcolor[HTML]{FFFFFF} & \cellcolor[HTML]{FFFFFF} & \cellcolor[HTML]{FFFFFF} & \cellcolor[HTML]{FFFFFF} & \cellcolor[HTML]{F5DCDC}1.15 & \cellcolor[HTML]{FFFFFF} & \cellcolor[HTML]{FFFFFF} & \cellcolor[HTML]{FFFFFF} & \cellcolor[HTML]{FFFFFF} & \cellcolor[HTML]{FFFFFF} & \cellcolor[HTML]{FFFFFF} & \cellcolor[HTML]{FFFFFF} \\
\textbf{lifestyle} & \cellcolor[HTML]{FFFFFF} & \cellcolor[HTML]{FFFFFF} & \cellcolor[HTML]{FFFFFF} & \cellcolor[HTML]{FFFFFF} & \cellcolor[HTML]{5B9BD5}0.64 & \cellcolor[HTML]{FFFFFF} & \cellcolor[HTML]{FFFFFF} & \cellcolor[HTML]{FFFFFF} & \cellcolor[HTML]{FFFFFF} & \cellcolor[HTML]{FFFFFF} & \cellcolor[HTML]{FFFFFF} & \cellcolor[HTML]{FFFFFF} \\
\textbf{memory} & \cellcolor[HTML]{FFFFFF} & \cellcolor[HTML]{FFFFFF} & \cellcolor[HTML]{5B9BD5}0.38 & \cellcolor[HTML]{FFFFFF} & \cellcolor[HTML]{FFFFFF} & \cellcolor[HTML]{FFFFFF} & \cellcolor[HTML]{FFFFFF} & \cellcolor[HTML]{FFFFFF} & \cellcolor[HTML]{FFFFFF} & \cellcolor[HTML]{FFFFFF} & \cellcolor[HTML]{5B9BD5}0.40 & \cellcolor[HTML]{FFFFFF} \\
\textbf{mental} & \cellcolor[HTML]{FFFFFF} & \cellcolor[HTML]{FFFFFF} & \cellcolor[HTML]{FFFFFF} & \cellcolor[HTML]{FFFFFF} & \cellcolor[HTML]{FFFFFF} & \cellcolor[HTML]{FFFFFF} & \cellcolor[HTML]{FFFFFF} & \cellcolor[HTML]{FFFFFF} & \cellcolor[HTML]{E6B5B6}1.33 & \cellcolor[HTML]{FFFFFF} & \cellcolor[HTML]{FFFFFF} & \cellcolor[HTML]{FFFFFF} \\
\textbf{money} & \cellcolor[HTML]{FFFFFF} & \cellcolor[HTML]{FFFFFF} & \cellcolor[HTML]{FFFFFF} & \cellcolor[HTML]{FFFFFF} & \cellcolor[HTML]{FFFFFF} & \cellcolor[HTML]{FFFFFF} & \cellcolor[HTML]{FFFFFF} & \cellcolor[HTML]{FFFFFF} & \cellcolor[HTML]{FFFFFF} & \cellcolor[HTML]{FFFFFF} & \cellcolor[HTML]{5B9BD5}0.57 & \cellcolor[HTML]{FFFFFF} \\
\textbf{moral} & \cellcolor[HTML]{FFFFFF} & \cellcolor[HTML]{FFFFFF} & \cellcolor[HTML]{FFFFFF} & \cellcolor[HTML]{FFFFFF} & \cellcolor[HTML]{FFFFFF} & \cellcolor[HTML]{FFFFFF} & \cellcolor[HTML]{FFFFFF} & \cellcolor[HTML]{FFFFFF} & \cellcolor[HTML]{FFFFFF} & \cellcolor[HTML]{FFFFFF} & \cellcolor[HTML]{5B9BD5}0.42 & \cellcolor[HTML]{FFFFFF} \\
\textbf{motion} & \cellcolor[HTML]{FFFFFF} & \cellcolor[HTML]{FFFFFF} & \cellcolor[HTML]{FFFFFF} & \cellcolor[HTML]{FFFFFF} & \cellcolor[HTML]{FFFFFF} & \cellcolor[HTML]{FFFFFF} & \cellcolor[HTML]{FFFFFF} & \cellcolor[HTML]{FFFFFF} & \cellcolor[HTML]{5B9BD5}0.64 & \cellcolor[HTML]{FFFFFF} & \cellcolor[HTML]{FFFFFF} & \cellcolor[HTML]{FFFFFF} \\
\textbf{perception} & \cellcolor[HTML]{FFFFFF} & \cellcolor[HTML]{DDEBF7}0.80 & \cellcolor[HTML]{FFFFFF} & \cellcolor[HTML]{FFFFFF} & \cellcolor[HTML]{DDEBF7}0.94 & \cellcolor[HTML]{FFFFFF} & \cellcolor[HTML]{FFFFFF} & \cellcolor[HTML]{DDEBF7}0.72 & \cellcolor[HTML]{FFFFFF} & \cellcolor[HTML]{FFFFFF} & \cellcolor[HTML]{FFFFFF} & \cellcolor[HTML]{FFFFFF} \\
\textbf{physical} & \cellcolor[HTML]{FFFFFF} & \cellcolor[HTML]{FFFFFF} & \cellcolor[HTML]{F5DCDC}1.16 & \cellcolor[HTML]{F5DCDC}1.09 & \cellcolor[HTML]{E6B5B6}1.45 & \cellcolor[HTML]{E6B5B6}1.27 & \cellcolor[HTML]{FFFFFF} & \cellcolor[HTML]{FFFFFF} & \cellcolor[HTML]{E6B5B6}1.44 & \cellcolor[HTML]{FFFFFF} & \cellcolor[HTML]{F5DCDC}1.13 & \cellcolor[HTML]{FFFFFF} \\
\textbf{polite} & \cellcolor[HTML]{FFFFFF} & \cellcolor[HTML]{D65C5E}2.39 & \cellcolor[HTML]{D65C5E}2.23 & \cellcolor[HTML]{D65C5E}2.70 & \cellcolor[HTML]{D65C5E}7.61 & \cellcolor[HTML]{D65C5E}5.80 & \cellcolor[HTML]{FFFFFF} & \cellcolor[HTML]{FFFFFF} & \cellcolor[HTML]{D65C5E}4.75 & \cellcolor[HTML]{FFFFFF} & \cellcolor[HTML]{E6B5B6}1.51 & \cellcolor[HTML]{D65C5E}1.72 \\
\textbf{ppron} & \cellcolor[HTML]{FFFFFF} & \cellcolor[HTML]{FFFFFF} & \cellcolor[HTML]{FFFFFF} & \cellcolor[HTML]{F5DCDC}1.14 & \cellcolor[HTML]{F5DCDC}1.14 & \cellcolor[HTML]{FFFFFF} & \cellcolor[HTML]{FFFFFF} & \cellcolor[HTML]{FFFFFF} & \cellcolor[HTML]{F5DCDC}1.14 & \cellcolor[HTML]{FFFFFF} & \cellcolor[HTML]{FFFFFF} & \cellcolor[HTML]{FFFFFF} \\
\textbf{pronoun} & \cellcolor[HTML]{FFFFFF} & \cellcolor[HTML]{FFFFFF} & \cellcolor[HTML]{FFFFFF} & \cellcolor[HTML]{F5DCDC}1.11 & \cellcolor[HTML]{FFFFFF} & \cellcolor[HTML]{FFFFFF} & \cellcolor[HTML]{FFFFFF} & \cellcolor[HTML]{FFFFFF} & \cellcolor[HTML]{FFFFFF} & \cellcolor[HTML]{FFFFFF} & \cellcolor[HTML]{F5DCDC}1.07 & \cellcolor[HTML]{FFFFFF} \\
\textbf{prosocial} & \cellcolor[HTML]{FFFFFF} & \cellcolor[HTML]{D65C5E}2.54 & \cellcolor[HTML]{DDEBF7}1.00 & \cellcolor[HTML]{F5DCDC}1.11 & \cellcolor[HTML]{5B9BD5}0.47 & \cellcolor[HTML]{E6B5B6}1.33 & \cellcolor[HTML]{FFFFFF} & \cellcolor[HTML]{FFFFFF} & \cellcolor[HTML]{DDEBF7}0.83 & \cellcolor[HTML]{FFFFFF} & \cellcolor[HTML]{F5DCDC}1.24 & \cellcolor[HTML]{FFFFFF} \\
\textbf{Qmark} & \cellcolor[HTML]{D65C5E}6.39 & \cellcolor[HTML]{D65C5E}5.19 & \cellcolor[HTML]{F5DCDC}1.15 & \cellcolor[HTML]{E6B5B6}1.28 & \cellcolor[HTML]{DDEBF7}0.85 & \cellcolor[HTML]{DDEBF7}0.94 & \cellcolor[HTML]{D65C5E}2.79 & \cellcolor[HTML]{D65C5E}5.03 & \cellcolor[HTML]{DDEBF7}0.86 & \cellcolor[HTML]{D65C5E}3.48 & \cellcolor[HTML]{D65C5E}2.70 & \cellcolor[HTML]{FFFFFF} \\
\textbf{relig} & \cellcolor[HTML]{FFFFFF} & \cellcolor[HTML]{FFFFFF} & \cellcolor[HTML]{D65C5E}2.76 & \cellcolor[HTML]{FFFFFF} & \cellcolor[HTML]{FFFFFF} & \cellcolor[HTML]{FFFFFF} & \cellcolor[HTML]{FFFFFF} & \cellcolor[HTML]{FFFFFF} & \cellcolor[HTML]{FFFFFF} & \cellcolor[HTML]{FFFFFF} & \cellcolor[HTML]{F5DCDC}1.14 & \cellcolor[HTML]{FFFFFF} \\
\textbf{reward} & \cellcolor[HTML]{FFFFFF} & \cellcolor[HTML]{FFFFFF} & \cellcolor[HTML]{FFFFFF} & \cellcolor[HTML]{FFFFFF} & \cellcolor[HTML]{FFFFFF} & \cellcolor[HTML]{FFFFFF} & \cellcolor[HTML]{FFFFFF} & \cellcolor[HTML]{FFFFFF} & \cellcolor[HTML]{FFFFFF} & \cellcolor[HTML]{FFFFFF} & \cellcolor[HTML]{5B9BD5}0.39 & \cellcolor[HTML]{D65C5E}5.71 \\
\textbf{risk} & \cellcolor[HTML]{FFFFFF} & \cellcolor[HTML]{FFFFFF} & \cellcolor[HTML]{FFFFFF} & \cellcolor[HTML]{FFFFFF} & \cellcolor[HTML]{D65C5E}1.98 & \cellcolor[HTML]{FFFFFF} & \cellcolor[HTML]{FFFFFF} & \cellcolor[HTML]{FFFFFF} & \cellcolor[HTML]{D65C5E}1.98 & \cellcolor[HTML]{FFFFFF} & \cellcolor[HTML]{FFFFFF} & \cellcolor[HTML]{FFFFFF} \\
\textbf{sexual} & \cellcolor[HTML]{FFFFFF} & \cellcolor[HTML]{FFFFFF} & \cellcolor[HTML]{FFFFFF} & \cellcolor[HTML]{FFFFFF} & \cellcolor[HTML]{FFFFFF} & \cellcolor[HTML]{FFFFFF} & \cellcolor[HTML]{FFFFFF} & \cellcolor[HTML]{FFFFFF} & \cellcolor[HTML]{FFFFFF} & \cellcolor[HTML]{FFFFFF} & \cellcolor[HTML]{5B9BD5}0.62 & \cellcolor[HTML]{FFFFFF} \\
\textbf{shehe} & \cellcolor[HTML]{FFFFFF} & \cellcolor[HTML]{5B9BD5}0.55 & \cellcolor[HTML]{FFFFFF} & \cellcolor[HTML]{FFFFFF} & \cellcolor[HTML]{FFFFFF} & \cellcolor[HTML]{FFFFFF} & \cellcolor[HTML]{FFFFFF} & \cellcolor[HTML]{5B9BD5}0.41 & \cellcolor[HTML]{FFFFFF} & \cellcolor[HTML]{FFFFFF} & \cellcolor[HTML]{DDEBF7}0.77 & \cellcolor[HTML]{FFFFFF} \\
\textbf{socbehav} & \cellcolor[HTML]{FFFFFF} & \cellcolor[HTML]{FFFFFF} & \cellcolor[HTML]{FFFFFF} & \cellcolor[HTML]{DDEBF7}0.92 & \cellcolor[HTML]{DDEBF7}0.70 & \cellcolor[HTML]{FFFFFF} & \cellcolor[HTML]{FFFFFF} & \cellcolor[HTML]{FFFFFF} & \cellcolor[HTML]{DDEBF7}0.77 & \cellcolor[HTML]{FFFFFF} & \cellcolor[HTML]{DDEBF7}0.99 & \cellcolor[HTML]{FFFFFF} \\
\textbf{social} & \cellcolor[HTML]{FFFFFF} & \cellcolor[HTML]{FFFFFF} & \cellcolor[HTML]{FFFFFF} & \cellcolor[HTML]{DDEBF7}0.98 & \cellcolor[HTML]{FFFFFF} & \cellcolor[HTML]{FFFFFF} & \cellcolor[HTML]{FFFFFF} & \cellcolor[HTML]{DDEBF7}0.70 & \cellcolor[HTML]{FFFFFF} & \cellcolor[HTML]{FFFFFF} & \cellcolor[HTML]{FFFFFF} & \cellcolor[HTML]{FFFFFF} \\
\textbf{socrefs} & \cellcolor[HTML]{FFFFFF} & \cellcolor[HTML]{FFFFFF} & \cellcolor[HTML]{FFFFFF} & \cellcolor[HTML]{F5DCDC}1.07 & \cellcolor[HTML]{F5DCDC}1.03 & \cellcolor[HTML]{FFFFFF} & \cellcolor[HTML]{FFFFFF} & \cellcolor[HTML]{5B9BD5}0.67 & \cellcolor[HTML]{FFFFFF} & \cellcolor[HTML]{FFFFFF} & \cellcolor[HTML]{FFFFFF} & \cellcolor[HTML]{FFFFFF} \\
\textbf{space} & \cellcolor[HTML]{FFFFFF} & \cellcolor[HTML]{FFFFFF} & \cellcolor[HTML]{FFFFFF} & \cellcolor[HTML]{FFFFFF} & \cellcolor[HTML]{F5DCDC}1.04 & \cellcolor[HTML]{FFFFFF} & \cellcolor[HTML]{FFFFFF} & \cellcolor[HTML]{DDEBF7}0.73 & \cellcolor[HTML]{DDEBF7}0.95 & \cellcolor[HTML]{FFFFFF} & \cellcolor[HTML]{FFFFFF} & \cellcolor[HTML]{FFFFFF} \\
\textbf{substances} & \cellcolor[HTML]{FFFFFF} & \cellcolor[HTML]{FFFFFF} & \cellcolor[HTML]{E6B5B6}1.27 & \cellcolor[HTML]{E6B5B6}1.33 & \cellcolor[HTML]{D65C5E}2.05 & \cellcolor[HTML]{FFFFFF} & \cellcolor[HTML]{FFFFFF} & \cellcolor[HTML]{FFFFFF} & \cellcolor[HTML]{D65C5E}2.65 & \cellcolor[HTML]{FFFFFF} & \cellcolor[HTML]{E6B5B6}1.28 & \cellcolor[HTML]{FFFFFF} \\
\textbf{swear} & \cellcolor[HTML]{FFFFFF} & \cellcolor[HTML]{FFFFFF} & \cellcolor[HTML]{FFFFFF} & \cellcolor[HTML]{FFFFFF} & \cellcolor[HTML]{FFFFFF} & \cellcolor[HTML]{FFFFFF} & \cellcolor[HTML]{FFFFFF} & \cellcolor[HTML]{FFFFFF} & \cellcolor[HTML]{FFFFFF} & \cellcolor[HTML]{FFFFFF} & \cellcolor[HTML]{5B9BD5}0.62 & \cellcolor[HTML]{FFFFFF} \\
\textbf{tentat} & \cellcolor[HTML]{FFFFFF} & \cellcolor[HTML]{D65C5E}1.73 & \cellcolor[HTML]{D65C5E}1.97 & \cellcolor[HTML]{D65C5E}2.14 & \cellcolor[HTML]{D65C5E}2.24 & \cellcolor[HTML]{E6B5B6}1.44 & \cellcolor[HTML]{FFFFFF} & \cellcolor[HTML]{FFFFFF} & \cellcolor[HTML]{D65C5E}2.21 & \cellcolor[HTML]{FFFFFF} & \cellcolor[HTML]{D65C5E}2.08 & \cellcolor[HTML]{FFFFFF} \\
\textbf{time} & \cellcolor[HTML]{FFFFFF} & \cellcolor[HTML]{FFFFFF} & \cellcolor[HTML]{FFFFFF} & \cellcolor[HTML]{DDEBF7}1.00 & \cellcolor[HTML]{F5DCDC}1.17 & \cellcolor[HTML]{FFFFFF} & \cellcolor[HTML]{FFFFFF} & \cellcolor[HTML]{FFFFFF} & \cellcolor[HTML]{F5DCDC}1.18 & \cellcolor[HTML]{FFFFFF} & \cellcolor[HTML]{FFFFFF} & \cellcolor[HTML]{FFFFFF} \\
\textbf{tone} & \cellcolor[HTML]{FFFFFF} & \cellcolor[HTML]{FFFFFF} & \cellcolor[HTML]{DDEBF7}0.85 & \cellcolor[HTML]{FFFFFF} & \cellcolor[HTML]{FFFFFF} & \cellcolor[HTML]{FFFFFF} & \cellcolor[HTML]{FFFFFF} & \cellcolor[HTML]{FFFFFF} & \cellcolor[HTML]{FFFFFF} & \cellcolor[HTML]{FFFFFF} & \cellcolor[HTML]{FFFFFF} & \cellcolor[HTML]{FFFFFF} \\
\textbf{tone\_neg} & \cellcolor[HTML]{FFFFFF} & \cellcolor[HTML]{5B9BD5}0.67 & \cellcolor[HTML]{FFFFFF} & \cellcolor[HTML]{FFFFFF} & \cellcolor[HTML]{F5DCDC}1.10 & \cellcolor[HTML]{FFFFFF} & \cellcolor[HTML]{FFFFFF} & \cellcolor[HTML]{FFFFFF} & \cellcolor[HTML]{F5DCDC}1.24 & \cellcolor[HTML]{FFFFFF} & \cellcolor[HTML]{FFFFFF} & \cellcolor[HTML]{FFFFFF} \\
\textbf{tone\_pos} & \cellcolor[HTML]{FFFFFF} & \cellcolor[HTML]{FFFFFF} & \cellcolor[HTML]{FFFFFF} & \cellcolor[HTML]{FFFFFF} & \cellcolor[HTML]{FFFFFF} & \cellcolor[HTML]{FFFFFF} & \cellcolor[HTML]{FFFFFF} & \cellcolor[HTML]{FFFFFF} & \cellcolor[HTML]{FFFFFF} & \cellcolor[HTML]{FFFFFF} & \cellcolor[HTML]{5B9BD5}0.58 & \cellcolor[HTML]{FFFFFF} \\
\textbf{visual} & \cellcolor[HTML]{FFFFFF} & \cellcolor[HTML]{FFFFFF} & \cellcolor[HTML]{FFFFFF} & \cellcolor[HTML]{FFFFFF} & \cellcolor[HTML]{FFFFFF} & \cellcolor[HTML]{FFFFFF} & \cellcolor[HTML]{FFFFFF} & \cellcolor[HTML]{5B9BD5}0.48 & \cellcolor[HTML]{FFFFFF} & \cellcolor[HTML]{FFFFFF} & \cellcolor[HTML]{DDEBF7}0.76 & \cellcolor[HTML]{FFFFFF} \\
\textbf{we} & \cellcolor[HTML]{FFFFFF} & \cellcolor[HTML]{FFFFFF} & \cellcolor[HTML]{5B9BD5}0.37 & \cellcolor[HTML]{FFFFFF} & \cellcolor[HTML]{FFFFFF} & \cellcolor[HTML]{FFFFFF} & \cellcolor[HTML]{FFFFFF} & \cellcolor[HTML]{5B9BD5}0.09 & \cellcolor[HTML]{5B9BD5}0.09 & \cellcolor[HTML]{FFFFFF} & \cellcolor[HTML]{5B9BD5}0.49 & \cellcolor[HTML]{FFFFFF} \\
\textbf{wellness} & \cellcolor[HTML]{FFFFFF} & \cellcolor[HTML]{FFFFFF} & \cellcolor[HTML]{FFFFFF} & \cellcolor[HTML]{FFFFFF} & \cellcolor[HTML]{FFFFFF} & \cellcolor[HTML]{FFFFFF} & \cellcolor[HTML]{FFFFFF} & \cellcolor[HTML]{FFFFFF} & \cellcolor[HTML]{FFFFFF} & \cellcolor[HTML]{FFFFFF} & \cellcolor[HTML]{5B9BD5}0.30 & \cellcolor[HTML]{D65C5E}3.74 \\
\textbf{you} & \cellcolor[HTML]{FFFFFF} & \cellcolor[HTML]{FFFFFF} & \cellcolor[HTML]{FFFFFF} & \cellcolor[HTML]{FFFFFF} & \cellcolor[HTML]{E6B5B6}1.34 & \cellcolor[HTML]{FFFFFF} & \cellcolor[HTML]{FFFFFF} & \cellcolor[HTML]{FFFFFF} & \cellcolor[HTML]{FFFFFF} & \cellcolor[HTML]{FFFFFF} & \cellcolor[HTML]{FFFFFF} & \cellcolor[HTML]{FFFFFF}

\\
\bottomrule
\end{tabular}%
}
\caption{Summary of statistically significant LIWC categories using Mann-Whitney U test. $R_I = \mu_{Inquisition}/\mu_{Disclosure}$ between the means of inquisition and disclosure-related posts, and $R_R = \mu_{Recovery}/\mu_{Dependency}$ between the means of recovery and dependency-related posts. Highlighted cells represent different $R$ with \colorbox[HTML]{5B9BD5}{dark blue} representing the lowest and \colorbox[HTML]{D65C5E}{dark red} representing the highest ratios. Empty cells denote non-significant
results.}
\label{tab:LIWC}
% \end{minipage} 
\end{table*}

\begin{table*}[h]
\centering
\resizebox{0.8\textwidth}{!}{%
\begin{tabular}{p{0.5\linewidth} >{\centering\arraybackslash}m{0.2\linewidth} >{\centering\arraybackslash}m{0.2\linewidth}}
\toprule
\multicolumn{1}{c}{\textbf{\texttt{Post}}} & \multicolumn{1}{c}{\textbf{\texttt{True Label}}} & \multicolumn{1}{c}{\textbf{\texttt{Predicted Label}}} \\\cmidrule(l){1-1}\cmidrule(l){2-3}
I was convinced the police were monitoring my computer and were about to arrest me. I bought it from a very reputable person—I'll spare the details—but he's been well-known with a good reputation for years. I'm still feeling a bit disoriented after that, and I'm going to die going to work tomorrow \textcolor{blue} {can someone please let me know what the fuck went wrong there} cheers
& Disclosure 
& Inquisition
\\\cmidrule(l){1-1}
I quit for about a week due to a scheduled blood test and surprisingly, I didn't feel as awful as expected. I don’t clearly remember that week, so I'm not certain how I felt. Later, I had my wisdom teeth removed and stopped again for a week before the procedure. During that time, the person I had been purchasing from unfortunately passed away. This was a big call and \textcolor{blue}{I have now been sober for about 3 months}, with all that information how bad was I again, 46 120 lb male
& Dependency
& Recovery
\\\cmidrule(l){1-1}
I want to care, but reality hits and I just want it to disappear. Music and weed, are the answer to a question I can't even ask. I know I can find satisfaction in that combination \textcolor{blue} {I got to quit},  bide me some time to figure out how to do it is that what I need to do
& Dependency &
 Recovery
\\\cmidrule(l){1-1}
Now I'm just inquiring because of \textcolor{blue}{my paranoia} about the prevalence of counterfeit medications these days. Are people really faking Etizest1 blister packs and boxes with no active ingredients or different, underdosed chemicals? I doubt they'd invest the effort to do that, considering some pharmaceutical companies in India manufacture our medicine, so there's definitely some \textcolor{blue} {level of quality control}.
& Safety & Legality, Quality, Safety, Mental Health
\\\cmidrule(l){1-1}
So, my usual dealer was out of supply for a few weeks, so they introduced me and my group of meth-using friends to another dealer. Turns out he had \textcolor{blue} {fire product and he is also an awesome person.}
& Quality & Quality, Nurturant support \& Morality\\
\bottomrule
\end{tabular}
}
\caption{Comparison of True and Predicted labels in some misclassified posts.}
\label{tab:label_comparison}
\end{table*}

\end{document}